\documentclass[10pt,twocolumn,letterpaper]{article}

\makeatletter
\@namedef{ver@everyshi.sty}{}
\makeatother
\usepackage{tikz}

\newif\ifarxiv
\arxivtrue %

\ifarxiv
    \usepackage[pagenumbers]{cvpr} %
\else
    \usepackage[review]{cvpr}      %
\fi

\setcitestyle{square, comma, numbers,sort&compress, super}
\usepackage{enumitem}
\usepackage{algpseudocode}
\usepackage[font=small,labelfont=bf]{caption}
\usepackage{array}
\usepackage{multirow}
\usepackage{booktabs}
\usepackage{algorithm}
\usepackage{subcaption}
\usepackage[normalem]{ulem}
\usepackage{xparse}
\usepackage{pifont}
\usepackage{bm}
\usepackage{threeparttable}
\usepackage{etoolbox}

\usepackage{listings}
\usepackage{mwe} %
\usepackage{makecell}
\usepackage{color, colortbl}
\usepackage{tabularx}
\usepackage{pifont}%
\usepackage[accsupp]{axessibility}
\usepackage[symbol]{footmisc}

\usepackage[scaled=0.85]{DejaVuSansMono}

\definecolor{citecolor}{HTML}{0071bc}
\usepackage[breaklinks=true,bookmarks=false,colorlinks,bookmarks=false, citecolor=citecolor, pagebackref]{hyperref}

\usepackage{graphicx}
\usepackage{amsmath}
\usepackage{amssymb}
\usepackage{courier}
\usepackage{pgfplots}

\pgfplotsset{compat=1.16}
\usepgfplotslibrary{groupplots}
\usetikzlibrary{patterns}
\usetikzlibrary{plotmarks}

\newlength\savewidth\newcommand\shline{\noalign{\global\savewidth\arrayrulewidth
  \global\arrayrulewidth 1pt}\hline\noalign{\global\arrayrulewidth\savewidth}}

\newlength\thinwidth

\definecolor{Gray}{gray}{0.92}
\definecolor{DarkGray}{gray}{0.5}
\newcolumntype{x}{>{\columncolor{Gray}}c}
\newcolumntype{H}{>{\setbox0=\hbox\bgroup}c<{\egroup}@{}}
\definecolor{LightCyan}{rgb}{0.88,1,1}
\definecolor{altRowColor}{gray}{0.92}
\definecolor{highlightRowColor}{rgb}{0.9, 0.9, 1}

\newcommand{\grayrow}{\rowcolor{Gray}}

\definecolor{GrayNumber}{gray}{0.5}
\definecolor{GrayXMark}{gray}{0.7}
\newcommand{\cmark}{\ding{51}}%
\newcommand{\xmark}{ {\color{GrayXMark} \ding{55}} } %

\newcommand{\aim}{{\textsc{AIMv1}}\xspace}
\newcommand{\OURS}{{\textsc{AIMv2}}\xspace}
\newcommand{\Ours}{\OURS}

\newcommand{\tablestyle}[2]{\setlength{\tabcolsep}{#1}\renewcommand{\arraystretch}{#2}\centering\footnotesize}

\expandafter\def\expandafter\normalsize\expandafter{%
    \normalsize%
    \setlength\abovedisplayskip{2pt}%
    \setlength\belowdisplayskip{5pt}%
    \setlength\abovedisplayshortskip{-5pt}%
    \setlength\belowdisplayshortskip{2pt}%
}

\newcommand{\cpar}{\par\noindent\textbf}

\newcommand{\graytext}{\textcolor{gray}}

\usepackage[capitalize]{cleveref}
\crefname{section}{\S}{\S\S}
\crefname{subsection}{\S}{\S\S}
\crefformat{table}{Table~#2#1#3}
\crefformat{figure}{Figure~#2#1#3}
\crefformat{equation}{Eq~#2#1#3}

\def\paperID{6276} %
\def\confName{CVPR}
\def\confYear{2025}

\title{Multimodal Autoregressive Pre-training of Large Vision Encoders}

\author{
    \scalebox{0.78}{Enrico Fini$^{\star}$ \quad \quad Mustafa Shukor$^{\star \dagger}$ \quad \quad Xiujun Li \quad \quad Philipp Dufter \quad \quad Michal Klein} \\
    \scalebox{0.78}{David Haldimann \quad \quad Sai Aitharaju \quad \quad Victor G. Turrisi da Costa \quad \quad Louis Béthune \quad \quad Zhe Gan} \\
    \scalebox{0.78}{Alexander Toshev \quad \quad Marcin Eichner \quad \quad Moin Nabi \quad \quad Yinfei Yang\quad \quad Joshua Susskind \quad \quad Alaaeldin El-Nouby$^{\star}$} \\ [0.2cm]
    \scalebox{0.95}{Apple} \\
    \small \url{https://github.com/apple/ml-aim}
}

\begin{document}
\maketitle
 
\ifarxiv
    \def\thefootnote{$\star$}\footnotetext{Equal technical contribution.\quad$\dagger$ Work done during an Internship.} \def\thefootnote{\arabic{footnote}}
    \def\thefootnote{}\footnotetext{Correspondence:\texttt{\{alaaeldin\_ali,efini\}@apple.com}}\def\thefootnote{\arabic{footnote}}
\fi

\begin{abstract}
We introduce a novel method for pre-training of large-scale vision encoders.
Building on recent advancements in autoregressive pre-training of vision models,
we extend this framework to a multimodal setting, i.e., images and text. In this
paper, we present \Ours, a family of generalist vision encoders
characterized by a straightforward pre-training process, scalability, and
remarkable performance across a range of downstream tasks. This is achieved by
pairing the vision encoder with a multimodal decoder that autoregressively
generates raw image patches and text tokens. Our encoders excel not only in
multimodal evaluations but also in vision benchmarks such as
localization, grounding, and classification. Notably, our \Ours-3B encoder
achieves 89.5\% accuracy on ImageNet-1k with a frozen trunk. Furthermore, \Ours
consistently outperforms state-of-the-art contrastive models (\eg, CLIP, SigLIP)
in multimodal image understanding across diverse settings.
\end{abstract}

\vspace{-5mm}
\section{Introduction}

\begin{figure*}[h!]
    \begin{minipage}[t]{0.5\linewidth}
    \vspace{0pt}
        \includegraphics[width=0.98\linewidth]{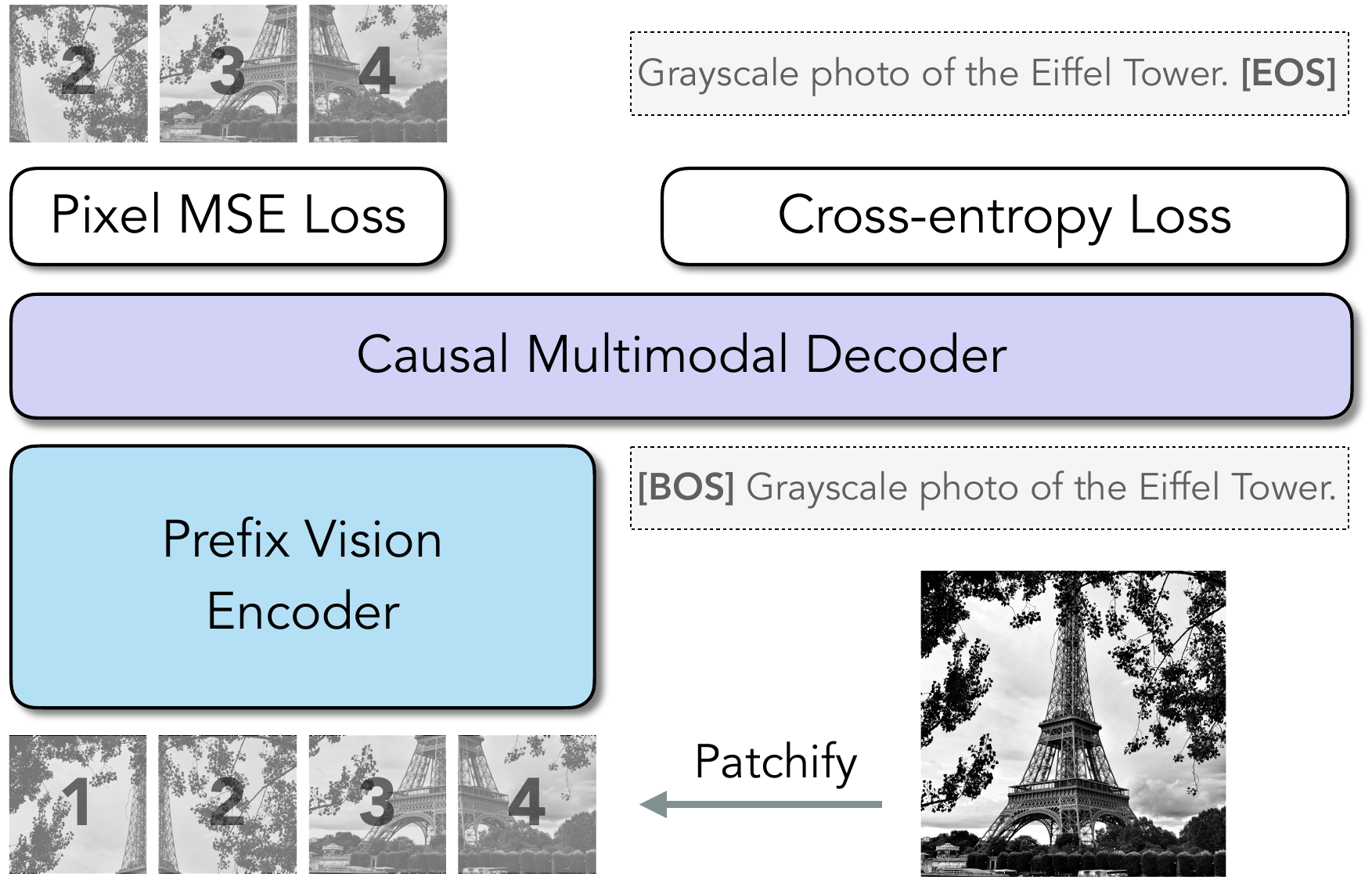}
    \end{minipage}
    \begin{minipage}[t]{0.5\linewidth}
        \vspace{-8pt}
        \begin{algorithm}[H]
            \caption*{\small{Pseudo-code for \OURS pre-training}}
            \label{algo:aimv2}
             \definecolor{codeblue}{rgb}{0.25,0.5,0.5}
             \definecolor{codekw}{rgb}{0.85, 0.18, 0.50}
             \lstset{
               basicstyle=\linespread{0.5}\fontsize{6.5pt}{6.5pt}\ttfamily,
               commentstyle=\fontsize{6pt}{6pt}\color{codeblue},
               keywordstyle=\fontsize{6pt}{6pt}\color{codekw},
             }
         \begin{lstlisting}[language=python, breaklines=false]
# img, cap: input image patches, caption
# I, T: number of patches, number of text tokens
# f_enc, f_dec: Transformer Encoder, Transformer Decoder

def forward(img, cap):
    # prepare targets by shifting the input
    pixel_target, cap_target = shift_left(img), shift_left(cap)

    # sample a prefix len and build attn mask
    attn_mask = build_attn_mask(prefix_len=randint(0, I-1))

    # extract image features
    img_feats = f_enc(I, attn_mask)

    # concatenate image features and text embedding
    mm_input = concat(img_feats, cap)

    # decode image patches and caption
    mm_out = f_dec(mm_input, is_causal=True)

    # compute loss
    pixel_loss = normalized_mse_loss(mm_out[:I], pixel_target)
    cap_loss = cross_entropy_loss(mm_out[-T:], cap_target)
    loss = cap_loss + alpha * pixel_loss
    return loss
            \end{lstlisting}  
         \end{algorithm}     
    \end{minipage}
    \captionof{figure}{\textbf{\Ours pre-training Overview.} (Left) Image
    patches are processed by a vision encoder trained with prefix
    attention~\citep{raffel2020exploring,el2024scalable}. The resulting visual
    representations are concatenated with the text embeddings of their
    corresponding captions. This combined multimodal sequence is then processed
    by a joint decoder. The model is pre-trained to autoregressively reconstruct
    the shifted input. (Right) Pseudocode for the forward pass during \Ours
    pre-training. The pre-training process of \Ours is straightforward to
    implement, resembling that of AIM and LLMs as it relies solely on a simple
    autoregressive objective.\label{fig:aimv2_overview}}
    \vspace{-5mm}
\end{figure*}

Research on pre-training of vision models has evolved significantly over time.
Initially, specialist models were designed to maximize performance on specific
tasks~\cite{he2016deep,he2017mask,carion2020end,feichtenhofer2018slowfast,karpathy2015deep,tolias2015particular}.
Gradually, general-purpose models emerged that can be deployed for a number of
pre-defined downstream tasks with minimal
adaptation~\cite{oquab2023dinov2,radford2021learning,jia2021scaling,zhai2023sigmoid}.
However, the remarkable success of Large Language Models
(LLMs)~\cite{touvron2023llama,reid2024gemini,achiam2023gpt,bai2023qwen} has
introduced new paradigms for utilizing vision
models~\cite{alayrac2022flamingo,mckinzie2024mm1,liu2024improvedllava,tong2024cambrian}.
Unlike the rigid predefined settings where vision models were previously
employed, LLMs enable more effective exploration of the pre-trained model
capabilities.  This shift demands rethinking pre-training methods for vision
models.

Generative pre-training is the dominant paradigm for language
modeling~\cite{radford2018improving,radford2019language,chowdhery2022palm} and
has shown remarkable performance and
scalability~\cite{hoffmann2022training,kaplan2020scaling}. Generative
pre-training has been extensively explored in computer
vision~\cite{he2021masked,bao2021beit,el2024scalable,doersch2015unsupervised,tschannen2024image},
but its performance still lags behind that of discriminative
methods~\cite{radford2021learning,oquab2023dinov2,zhai2023sigmoid,zhou2021ibot}.
For instance, a formulation highly reminiscent of LLMs pre-training was proposed
by~\citet{el2024scalable} and demonstrated encouraging scaling properties.
However, it requires much higher capacity models to match the performance of its
discriminative counterparts.  In contrast, while contrastive techniques are
often more parameter efficient, they are notably challenging to train and scale.
Although significant progress has been made to mitigate these issues, there
remains a gap in developing methods that combine the simplicity and scalability
of generative pre-training with the parameter efficiency of discriminative
approaches.

In this paper, we introduce \Ours, a family of open vision models pre-trained to
autoregressively generate both image patches and text tokens. During
pre-training, \Ours uses a causal multimodal decoder that first regresses image
patches and then decodes text tokens in an autoregressive manner, as illustrated
in~\cref{fig:aimv2_overview}. Such a simple approach offers several advantages.
First, \Ours is straightforward to implement and train without requiring
excessively large batch sizes~\cite{radford2021learning,fang2023data} or
specialized inter-batch communication methods~\cite{zhai2023sigmoid}. Second,
the architecture and pre-training objectives of \Ours align well with
LLM-powered multimodal applications, enabling seamless integration. Finally,
\Ours extracts a training signal from every image patch and text token,
providing denser supervision compared to discriminative objectives.

Our \Ours models are strong generalists that exhibit remarkable performance
across various vision and multimodal tasks. In particular, \Ours performs
favorably on multimodal understanding benchmarks compared to state-of-the-art
vision-language pre-trained methods~\cite{zhai2023sigmoid,fang2023data}. It
outperforms DINOv2~\cite{oquab2023dinov2} on open-vocabulary object detection
and referring expression comprehension, and attains strong recognition
performance with a frozen trunk, outperforming a number of strong baselines.
Furthermore, \Ours enjoys strong scalability, similar to its language-only and
vision-only counterparts, improving consistently when scaling data or
parameters. Moreover, we demonstrate the compatibility of \Ours with several
modern tools, including support for native image resolution and adaptation to
zero-shot recognition~\cite{zhai2022lit}. We discuss related works in more
detail in~\cref{sec:related_works}.

\section{Approach}

\subsection{Pre-training}
\label{sec:pretraining_method}

\begin{figure*}[t!]
    \centering
    \captionsetup{type=figure}
    \begin{subfigure}[t]{0.48\linewidth}
        \definecolor{CustomA}{HTML}{f1c40f}
\definecolor{CustomB}{HTML}{d68910}
\definecolor{CustomC}{HTML}{17a589}
\definecolor{CustomD}{HTML}{2980b9}

\begin{tikzpicture}
    \begin{axis}[
        legend pos=north east,
        grid=both,
        grid style={line width=.1pt, draw=gray!10},
        major grid style={line width=.2pt,draw=gray!50},
        minor tick num=2,
        axis x line*=bottom,
        axis y line*=left,
        xtick={10, 20, 30, 40, 50, 60},
        xticklabels={\texttt{1E9}, \texttt{2E9}, \texttt{3E9}, \texttt{4E9}, \texttt{5E9}, \texttt{6E9}},
        xmin=0,
        xmax=70,
        height=2.0in,
        width=1.05\linewidth,
        ylabel style={align=center, font=\footnotesize, yshift=-0.5ex},
        xlabel style={font=\footnotesize},
        ylabel={\footnotesize{Validation Cross-Entropy}},
        xlabel={\footnotesize{Samples Seen}},
        yticklabel style={font=\footnotesize},
        xticklabel style={font=\footnotesize},
        legend style={cells={align=left}, font=\footnotesize}, %
        legend cell align={left},
        mark options={solid},
    ]

    \addlegendimage{mark=none, only marks, mark size=0.0001pt}
    \addlegendentry{\hspace{-.6cm}\texttt{parameters}}

   \addlegendimage{mark=*, CustomD, mark options={solid}, line width=1.5pt, mark size=2pt}
    \addlegendentry{3e8}

   \addlegendimage{mark=*, CustomC, mark options={solid}, line width=1.5pt, mark size=2pt}
    \addlegendentry{6e8}

   \addlegendimage{mark=*, CustomB, mark options={solid}, line width=1.5pt, mark size=2pt}
    \addlegendentry{1e9}
    
    \addlegendimage{mark=*, CustomA, mark options={solid}, line width=1.5pt, mark size=2pt}
    \addlegendentry{3e9}

    \addplot[mark=oplus*, CustomD, line width=1.5pt, mark size=2pt] plot coordinates {
        (4.92,  2.844)
        (9.83,  2.764)
        (14.7,  2.733)
        (19.7,  2.711)
        (29.5,  2.683)
        (39.3,  2.643)
        (63.9,  2.641)
    };

    \addplot[mark=oplus*, CustomC, line width=1.5pt, mark size=2pt] plot coordinates {
        (4.92,  2.812)
        (9.83,  2.712)
        (14.7,  2.673)
        (19.7,  2.66)
        (29.5,  2.631)
        (39.3,  2.627)
        (63.9,   2.607)
    };

    \addplot[mark=oplus*, CustomB, line width=1.5pt, mark size=2pt] plot coordinates {
        (4.92,  2.779)
        (9.83,  2.688)
        (14.7,  2.664)
        (19.7,  2.643)
        (29.5,  2.621)
        (39.3, 2.589)
        (63.9, 2.589)
    };

    \addplot[mark=oplus*, CustomA, line width=1.5pt, mark size=2pt] plot coordinates {
        (4.92, 2.762)
        (9.83, 2.676)
        (14.7, 2.659)
        (19.7, 2.639)
        (29.5, 2.604)
        (39.3, 2.578)
        (63.9, 2.555)
    };

    \end{axis}
\end{tikzpicture}
    \end{subfigure}
    \begin{subfigure}[t]{0.48\linewidth}
        \definecolor{CustomA}{HTML}{f1c40f}
\definecolor{CustomB}{HTML}{d68910}
\definecolor{CustomC}{HTML}{17a589}
\definecolor{CustomD}{HTML}{2980b9}

\begin{tikzpicture}
    \begin{axis}[
        legend pos=north east,
        grid=both,
        grid style={line width=.1pt, draw=gray!10},
        major grid style={line width=.2pt,draw=gray!50},
        minor tick num=2,
        axis x line*=bottom,
        axis y line*=left,
        xtick={5, 15, 50, 150, 450},
        xticklabels={\texttt{5E19}, \texttt{15E19}, \texttt{5E20}, \texttt{15e21}, \texttt{45E21}},
        xmin=3,
        xmax=480,
        height=2in,
        width=1.05\linewidth,
        ylabel style={align=center, font=\footnotesize, yshift=-0.5ex},
        xlabel style={font=\footnotesize},
        ylabel={\footnotesize{Validation Cross-Entropy}},
        xlabel={\footnotesize{FLOPs \scriptsize{(log scale)}}},
        yticklabel style={font=\footnotesize},
        xticklabel style={font=\footnotesize},
        legend style={cells={align=left}, font=\footnotesize}, %
        legend cell align={left},
        xmode=log, %
        mark options={solid},
    ]

    \addlegendimage{mark=none, only marks, mark size=0.0001pt}
    \addlegendentry{\hspace{-.6cm}\texttt{parameters}}

   \addlegendimage{mark=*, CustomD, mark options={solid}, line width=1.5pt, mark size=2pt}
    \addlegendentry{3e8}

   \addlegendimage{mark=*, CustomC, mark options={solid}, line width=1.5pt, mark size=2pt}
    \addlegendentry{6e8}

   \addlegendimage{mark=*, CustomB, mark options={solid}, line width=1.5pt, mark size=2pt}
    \addlegendentry{1e9}
    
    \addlegendimage{mark=*, CustomA, mark options={solid}, line width=1.5pt, mark size=2pt}
    \addlegendentry{3e9}

    \addplot[mark=oplus*, CustomD, line width=1.5pt, mark size=2pt] plot coordinates {
        (3.89,  2.844)
        (7.77,  2.764)
        (11.7,  2.733)
        (15.5,  2.711)
        (23.3,  2.683)
        (31.1,  2.643)
        (50.5,  2.641)
    };

    \addplot[mark=oplus*, CustomC, line width=1.5pt, mark size=2pt] plot coordinates {
        (8.57,  2.812)
        (17.1,  2.712)
        (25.7,  2.673)
        (34.2,  2.66)
        (51.4,  2.631)
        (68.5,  2.627)
        (111,   2.607)
    };

    \addplot[mark=oplus*, CustomB, line width=1.5pt, mark size=2pt] plot coordinates {
        (15.5,  2.779)
        (31.1,  2.688)
        (46.6,  2.664)
        (62.1,  2.643)
        (93.2,  2.621)
        (124.3, 2.589)
        (202, 2.589)
    };

    \addplot[mark=oplus*, CustomA, line width=1.5pt, mark size=2pt] plot coordinates {
        (34.2, 2.762)
        (68.4, 2.676)
        (103, 2.659)
        (137, 2.639)
        (205, 2.604)
        (274, 2.578)
        (445, 2.555)
    };

    \end{axis}
\end{tikzpicture}
    \end{subfigure}
    \vspace{-5mm}
    \caption{\textbf{Scaling properties of \Ours.} (Left) Given a fixed
    pre-training data size, increasing the number of parameters always leads to
    an improvement in the validation loss. (Right) The optimal model size
    varies based on the pre-training compute budget. Larger models perform worse
    than smaller ones when severely undertrained but improves consistently as
    the compute budget increases. This behavior is consistent with that reported
    by~\citet{hoffmann2022training} for text-only autoregressive
    models.\label{fig:scaling_laws}}
    \vspace{-3mm}
\end{figure*}
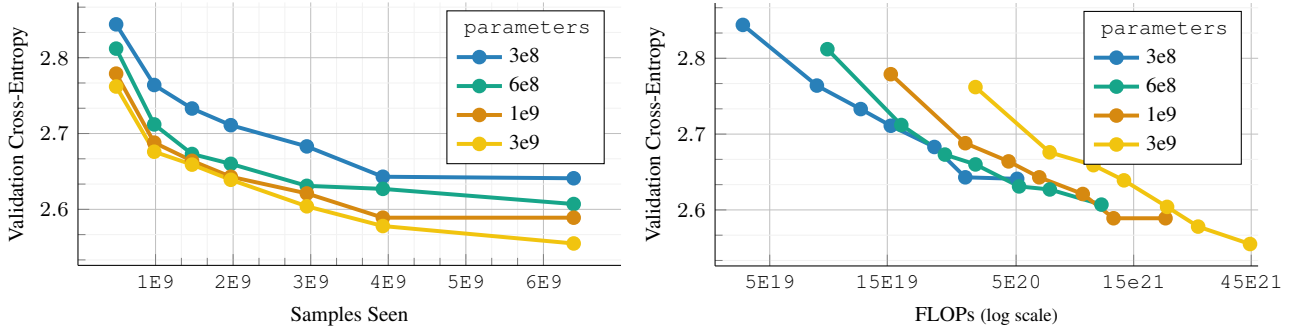

Our model extends the standard unimodal autoregressive framework to multimodal
settings that integrate both images and text into a unified sequence.
Specifically, an image $x$ is partitioned into $I$ non-overlapping patches
$x_i$, $i \in [1, I]$, forming a sequence of tokens. Similarly, a text sequence
is broken down into subwords $x_t$, $t \in [I, I+T]$. These sequences are then
concatenated, allowing text tokens to attend to image tokens. While both
concatenation directions (image $\rightarrow$ text and text $\rightarrow$ image)
are possible, we focus on training a strong vision encoder by always prepending
the image first, thereby enabling stronger conditioning on the visual features.
This results in a unified multimodal autoregressive modeling process, where the
sequence is factorizatized as follows:
\begin{align*}
    &P(S) = \prod_{j=1}^{I+T} P(S_j | S_{<j}),
\end{align*}
where $S_j$ represents the $j$-th token in the concatenated sequence of image
patches and text tokens, and $S_{<j}$ includes all preceding tokens. This
unified factorization allows the model to autoregressively predict the next
token in the sequence, regardless of what modality it is currently processing.
Our pre-training setup consists of a dedicated vision encoder that processes the
raw image patches, which are then passed to a multimodal decoder alongside the
embedded text tokens, as illustrated in~\cref{fig:aimv2_overview}. The decoder
subsequently performs next-token prediction on the combined sequence, following
the factorization above. To support the autoregressive generation process, the
vision encoder and multimodal decoder employ prefix and causal self-attention
operations, respectively.

\cpar{Objective function.} We define separate loss functions for the image and
text domains as follows:
\begin{align*}
    L_{\text{img}} = \frac{1}{I} \sum_{i=1}^{I} \| \hat{x}_i(x_{<i}; \theta) - x_i \|_2^2, \\
    L_{\text{text}} = - \frac{1}{T} \sum_{t=I+1}^{I+T} \log P(x_t | x_{<t}; \theta).
\end{align*}
The overall objective is to minimize \( L_{\text{text}} + \alpha *
L_{\text{img}} \) with respect to model parameters \( \theta \). For the text
domain, \( L_{\text{text}} \) is a standard cross-entropy loss that measures the
negative log-likelihood of the ground truth token at each step. For the image
domain, \( L_{\text{img}} \) is an \( \ell_2 \) pixel-level regression loss,
where the model's predicted patch \(\hat{x}_i(\theta)\) is compared to the true
patch \(x_i\). We normalize the image patches following~\citet{he2021masked}.
In practice, we use separate linear layers to map the final hidden state of the
multimodal decoder to the appropriate output dimensions for image patches and
vocabulary size for vision and language, respectively.

\subsection{Architecture}

For the vision encoder of \Ours, we adopt the Vision Transformer (ViT)
architecture~\cite{dosovitskiy2020image}. We train a series of vision encoders
ranging between 300M and 3B parameters. Detailed model specifications are
provided in \cref{tab:model_specs}.

\begin{table}[t]
    \centering
    \setlength{\tabcolsep}{11pt}
    \renewcommand{\arraystretch}{1}
    \resizebox{1\linewidth}{!}{
    \begin{tabular}{lccccc}
        model & \#params  & $d_{\text{enc}}$ & ${L_{\text{enc}}}$ &
        $d_{\text{dec}}$ & ${L_{\text{dec}}}$ \\
         \shline
         \Ours-L  & 0.3B & 1024  & \multirow{4}{*}{24} & \multirow{4}{*}{12}
         & \multirow{4}{*}{1024} \\
         \Ours-H  & 0.6B & 1536  & \\
         \Ours-1B & 1.2B & 2048  & \\ 
         \Ours-3B & 2.7B & 3072  &  \\ 
    \end{tabular}} \caption{\textbf{\Ours family of models.} We detail the
    architectural specifications of \Ours models including the embedding
    dimension $d$, number of layers ${L}$ for the vision encoder and the
    mutlimodal decoder, and the total number of encoder parameters.}
    \label{tab:model_specs}
\end{table}

\cpar{Prefix Attention.} Following \citet{el2024scalable}, we constrain the
self-attention mechanism within the vision encoder by applying a prefix
attention mask \cite{raffel2020exploring}. This strategy facilitates the use of
bidirectional attention during inference without additional tuning.
Specifically, we randomly sample the prefix length as \( M \sim \mathcal{U}\{1,
2, \dots, I - 1\} \). The pixel loss is computed exclusively for non-prefix
patches, defined as \( \{\, x_i \mid i > M \,\} \).

\cpar{SwiGLU and RMSNorm.} Our vision encoder and multimodal decoder incorporate
SwiGLU \cite{shazeer2020glu} as the feed-forward network (FFN) and replace all
normalization layers with RMSNorm \cite{zhang2019root}. These modifications
leverage the recent successes of SwiGLU and RMSNorm in language modeling
\cite{touvron2023llama,touvron2023llama2}.

\cpar{Multimodal Decoder.} We adopt a unified multimodal decoder that performs
autoregressive generation for both image and text modalities concurrently. Image
features and raw text tokens are each linearly projected and embedded into \(
\mathbb{R}^{d_{\text{dec}}} \). The decoder receives concatenated sequences of
image and text features as input and employs causal attention in the
self-attention operations. The outputs of the decoder are processed through two
separate linear heads—one for image tokens and another for text tokens—to
predict the next token in each modality respectively. We use the same decoder
capacity for all the \Ours variants.

\noindent  The optimization hyperparameters used during pre-training of all
\Ours models are outlined in~\cref{tab:pretrain_hparams}.
\subsection{Data}

We pre-train \Ours models using a combination of public and private datasets
containing paired images and text. We use the publicly available
DFN-2B~\cite{fang2023data} and COYO~\cite{byeon2022coyo} datasets, along with a
proprietary dataset of High Quality Image-Text Pairs (HQITP). In addition to
alt-text, we use synthetic captions following the approach
of~\citet{lai2024revisit}. Details regarding the datasets, including their sizes
and the sampling probabilities used for each dataset, are provided in
\cref{tab:pretraining_datasets}. Unless mentioned otherwise, all \Ours models
were pre-trained using 12 billion image-text samples.

\begin{table}[t!]
    \centering
    \setlength{\tabcolsep}{8pt}
    \renewcommand{\arraystretch}{1}
    \resizebox{1\linewidth}{!}{
    \begin{tabular}{lccrc}
        dataset & public & caption & \#images-text pairs & sampling prob. \\
         \shline
         \multirow{2}{*}{DFN~\citep{fang2023data}} & \cmark &
         \textit{alt-text} &1,901,228,573 & 30\% \\
          & \xmark & \textit{synthetic} & 3,802,457,146 & 30\% \\
          COYO~{\citep{byeon2022coyo}} & \cmark & \textit{alt-text}
          & 560,171,533 & 9\% \\
          \multirow{2}{*}{HQITP} & \xmark & \textit{alt-text}  & 564,623,839 &
          28\% \\
          & \xmark & \textit{synthetic}  & 431,506,953 & 3\% \\ 
    \end{tabular}} \caption{\textbf{Pre-training data mixture.} \Ours is
    pre-trained using a large-scale collection of image and text pairs. For the
    paired captions, we utilize alt-text as well as synthetic text generated
    from a pre-trained captioner. In this table we list the datasets as well the
    sampling probabilities we used for each data source. }
    \label{tab:pretraining_datasets}
\end{table}

\begin{figure*}[t!]
    \centering
    \begin{minipage}[t]{0.33\linewidth}
        \captionsetup{type=figure}
            \begin{tikzpicture}
    \begin{axis}[
        legend pos=south east,
        grid=both,
        grid style={line width=.1pt, draw=gray!10},
        major grid style={line width=.2pt,draw=gray!50},
        minor tick num=2,
        axis x line*=bottom,
        axis y line*=left,
        xtick={1, 2, 3, 4},
        xticklabels={L, H, 1B, 3B},
        ymin=85.6, ymax=90,
        xlabel={Model Size},
        ylabel={IN-1k Accuracy (\%)},
        ylabel style={align=center, font=\footnotesize, yshift=-0.5ex},
        xlabel style={font=\footnotesize},
        yticklabel style={font=\footnotesize},
        xticklabel style={font=\footnotesize},
        legend style={cells={align=left}, font=\footnotesize},
        legend cell align={left},
        mark options={solid},
        width=1.05\linewidth,
        height=2in,
    ]

    \definecolor{pastelBlue}{RGB}{173, 216, 230}
    \definecolor{pastelGreen}{RGB}{152, 251, 152}
    \definecolor{pastelRed}{RGB}{255, 182, 193}
    \definecolor{CustomA}{HTML}{f1c40f}
    \definecolor{CustomB}{HTML}{d68910}
    \definecolor{CustomC}{HTML}{17a589}
    \definecolor{CustomD}{HTML}{2980b9}
    
    \addplot[mark=oplus*, color=CustomD, line width=1.5pt, mark size=2pt] coordinates {
        (1, 86.6) (2, 87.5) (3, 88.1) (4, 88.5)
    };
    \addlegendentry{224px}

    \addplot[mark=oplus*, color=CustomC, line width=1.5pt, mark size=2pt] coordinates {
        (1, 87.6) (2, 88.2) (3, 88.7) (4, 89.2)
    };
    \addlegendentry{336px}

    \addplot[mark=oplus*, color=pastelRed!200, line width=1.5pt, mark size=2pt] coordinates {
        (1, 87.9) (2, 88.6) (3, 89.0) (4, 89.5)
    };
    \addlegendentry{448px}

    \end{axis}
\end{tikzpicture}
        \vspace{1.5mm}
        \caption{\textbf{Scaling capacity and resolution.} \Ours shows
        strong scalability with respect to model parameters,  measured in frozen-trunk top-1
        accuracy for IN-1k. This behavior is consistent when scaling image resolution.}
        \label{fig:scaling_model_res}
 \end{minipage}\hfill
    \begin{minipage}[t]{0.64\linewidth}
        \captionsetup{type=figure}
        \begin{subfigure}[t]{0.49\linewidth}
            \begin{tikzpicture}
    \begin{axis}[
        width=1.05\linewidth,
        height=2in,
        xlabel={Image-text Pairs},
        ylabel={IN-1k Accuracy (\%)},
        xtick={1,2,3,4},
        xticklabels={500M, 1B, 2B, 4B},
        ymin=84, ymax=87,
        grid=both,
        axis x line*=bottom,
        axis y line*=left,
        grid style={line width=.1pt, draw=gray!10},
        major grid style={line width=.2pt,draw=gray!50},
        tick align=outside,
        tick pos=left,
        axis line style={thin},
        major tick length=0.1cm,
        legend pos=south east,
        legend style={cells={align=left}, font=\footnotesize}, %
        legend cell align={left},
        legend image post style={scale=1.2},
        every axis label/.append style={font=\footnotesize},
        label style={font=\footnotesize},
        tick label style={font=\footnotesize},
        title style={font=\large, yshift=-1ex}
    ]

    \addplot[
        mark=*,
        mark size=1.7pt,
        line width=1pt,
        black
    ] coordinates {
        (1, 84.850) (2, 85.874) (3, 86.5479) (4, 86.877)
    };
    \addlegendentry{\Ours}

    \addplot[
        mark=square*,
        mark options={solid},
        mark size=1.4pt,
        line width=1pt,
        dashed,
        gray
    ] coordinates {
        (1, 84.242) (2, 85.568) (3, 86.2654) (4, 86.345)
    };
    \addlegendentry{Cap}

    \addplot [
        fill=gray, 
        fill opacity=0.3, 
        draw=none
    ] coordinates {
        (1, 84.850) (2, 85.874) (3, 86.5479) (4, 86.877)
    } -- 
    (axis cs: 4, 86.345) -- 
    (axis cs: 3, 86.2654) -- 
    (axis cs: 2, 85.568) -- 
    (axis cs: 1, 84.242) -- 
    cycle;

    \end{axis}
\end{tikzpicture}
        \end{subfigure}
        \hfill
        \begin{subfigure}[t]{0.49\linewidth}
            \begin{tikzpicture}
    \begin{axis}[
        width=1.05\linewidth,
        height=2in,
        xlabel={Model Size},
        ylabel={IN-1k Accuracy (\%)},
        xtick={1,2,3},
        xticklabels={L, H, 1B},
        ymin=85, ymax=87,
        grid=both,
        axis x line*=bottom,
        axis y line*=left,
        grid style={line width=.1pt, draw=gray!10},
        major grid style={line width=.2pt,draw=gray!50},
        tick align=outside,
        tick pos=left,
        axis line style={thin},
        major tick length=0.1cm,
        legend pos=south east,
        legend style={cells={align=left}, font=\footnotesize},
        legend cell align={left},
        legend image post style={scale=1.2},
        every axis label/.append style={font=\footnotesize},
        label style={font=\footnotesize},
        tick label style={font=\footnotesize},
        title style={font=\large, yshift=-1ex}
    ]

    \addplot[
        mark=*,
        mark size=1.7pt,
        line width=1pt,
        black
    ] coordinates {
        (1, 85.603) (2, 86.5479) (3, 86.92175)
    };
    \addlegendentry{\Ours}

    \addplot[
        mark=square*,
        mark options={solid},
        mark size=1.4pt,
        line width=1pt,
        dashed,
        gray
    ] coordinates {
        (1, 85.40325) (2, 86.288) (3, 86.69675)
    };
    \addlegendentry{Cap}

    \addplot [
        fill=gray, 
        fill opacity=0.3, 
        draw=none
    ] coordinates {
        (1, 85.603) (2, 86.5479) (3, 86.92175)
    } -- 
    (axis cs: 3, 86.69675) -- 
    (axis cs: 2, 86.288) -- 
    (axis cs: 1, 85.40325) -- 
    cycle;

    \end{axis}
\end{tikzpicture}
        \end{subfigure}       
        \caption{\textbf{\Ours \vs Captioning.} We investigate the role of the image-level objective in comparison to a
        captioning-only baseline, particularly as we scale data and model size. Our
        findings indicate that \Ours consistently outperforms the captioning
        baseline across all dataset and model sizes. Notably, \Ours exhibits fewer
        signs of saturation when scaling data compared to the
        captioning-only approach.}
        \label{fig:aimv2_vs_cap_plot}
    \end{minipage}
    \vspace{-5mm}
\end{figure*}

\subsection{Post-Training} 
\label{sec:post_training}

While the initial pre-training stage of \Ours yields highly performant models,
we explore methods to further enhance the capabilities through various
post-training strategies.

\cpar{High-resolution Adaptation.} In the initial pre-training stage, we use
image data with a fixed resolution of 224px. However, many downstream task, such
as detection, segmentation, and multimodal LLMs, benefit from models adapted to
handle higher resolution images. Therefore, we finetune \Ours models for 336 and
448 pixel resolutions. The high-resolution adaptation stage utilizes 2 billion
image-text pairs sampled from the same pool as the pre-training stage, except
that we do not use synthetic captions at this stage. Consistent with the
observations of~\citet{zhai2023sigmoid}, we find that weight decay of zero is
important for maintaining stable optimization. %

\cpar{Native Resolution Fine-tuning.} Training models for a dedicated resolution
and aspect ratio can be inflexible for many applications that require processing
images in their original shapes. Prior works such as
FlexiViT~\cite{beyer2023flexivit} and NaViT~\cite{dehghani2024patch} have
tackled these limitation. We adopt a different approach for training with
variable aspect ratios and resolutions. Specifically, we define $B_i$ as the
number of images in a mini-batch, $A_i$ as the number of patches per image, and
$C$ as the total number of image patches in the mini-batch. For a mini-batch
$i$, we randomly sample an area $A$ and resize the images to fit within this
area while maintaining their aspect ratios.\footnote{We use zero padding if the
image cannot be perfectly fitted into the desired area.} We then adjust the
mini-batch size $B_i$ such that $C = A_i \times B_i$. This strategy is analogous
to the approach proposed by~\citet{pouransari2024dataset} for training LLMs with
variable context lengths. Our implementation does not require heuristics for
sequence packing, attention masking, or custom pooling operations. We choose $A
= 2^{n}$, where $n$ is sampled from a truncated normal distribution
$\mathcal{N}(0, 1)$ within the range \([-1, 1]\) and linearly mapped to \([7,
12]\).

\section{Analysis}

One of the main advantages of \Ours is its simplicity; it is easy to implement
and scale. Therefore, we investigate the scaling properties of the \Ours family
of models.

\subsection{Scaling AIMv2}

First, we investigate the impact of scaling data size and model capacity on the
validation performance of \Ours. We fix the model size and vary the number of
samples seen during pre-training. This analysis is similar to
``\textit{Approach~1}'' in the study of \citet{hoffmann2022training}. The
results of this study are illustrated in~\cref{fig:scaling_laws}.

\cpar{Setup.} We train four model capacities, ranging from 300 million to 3
billion parameters, and vary the number of samples seen between 500 million to
6.4 billion image-text pairs. All models are trained to convergence with no
early stopping. To achieve this with minimal computational cost, we train a
single model for each capacity using 5 billion images with a half-cosine
learning rate schedule (\ie, the final learning rate is half the peak learning
rate). We select seven intermediate checkpoints from this run and apply a linear
cooldown to \(1 \times 10^{-6}\). The length of the cooldown stage is 20\% of
the initial pre-training stage.

\cpar{Results.} We observe a consistent improvement in performance with scaling
data or parameters. However, diminishing returns appear when scaling data for
the lower-capacity models. Additionally, we find that the optimal model size
changes as the compute budget varies. At smaller compute budgets,
larger-capacity models are undertrained and underperform compared to their
lower-capacity counterparts.

\begin{table*}[t!]
    \centering
    \setlength{\tabcolsep}{11pt}
    \renewcommand{\arraystretch}{1}
    \resizebox{1\linewidth}{!}{
    \begin{tabular}{llccccccccccccccc}
        model & architecture & \small{\rotatebox{90}{IN-1k}} & \small{\rotatebox{90}{iNAT-18}} & \small{\rotatebox{90}{Cifar10}}  & \small{\rotatebox{90}{Cifar100}} & \small{\rotatebox{90}{Food101}} & \small{\rotatebox{90}{DTD}} & \small{\rotatebox{90}{Pets}} & \small{\rotatebox{90}{Cars}} & \small{\rotatebox{90}{CAM17}} & \small{\rotatebox{90}{PCAM}} & \small{\rotatebox{90}{RxRx1}}  & \small{\rotatebox{90}{EuroSAT}} & \small{\rotatebox{90}{fMoW}} & \small{\rotatebox{90}{Infographic}}\\
         \shline
        \multirow{1}{*}{MAE~\cite{he2021masked}} & ViT-2B/14
        & 82.2 %
        & 70.8 %
        & 97.5
        & 87.3
        & 93.4 %
        & 81.2 %
        & 95.1 %
        & 94.9 %
        & 94.4    %
        & 90.3    %
        & 7.3    %
        & 98.2 %
        & 60.1 %
        & 50.2
        \\
         \multirow{2}{*}{\aim~\cite{el2024scalable}}  & ViT-H/14 
          & 78.5 %
          & 64.0 %
          & 97.2
          & 86.8
          & 90.1 %
          & 80.1 %
          & 93.0 %
          & 93.0 %
          & 94.3  %
          & 90.0 %
          & 7.8 %
          & 98.4 %
          & 58.3 %
          & 45.2
          \\

          & ViT-7B/14 &
           84.0 %
          & 75.5 %
          & 98.9 
          & 91.8
          & 94.1 %
          & 85.6 %
          & 95.4 %
          & 95.0 %
          & 94.2  %
          & 90.5 %
          & 8.4 %
          & 98.5 %
          & 63.5 %
          & 57.7
           \\	
         DINOv2~\cite{oquab2023dinov2} & ViT-g/14 %
         &    87.2
         &    83.0
         &    99.7
         &    95.6
         &    96.0
         &    86.9
         &    96.8
         &    94.9
         &    95.8
         &    90.1
         &    9.0
         &    98.8
         &    65.5
         &    59.4
         \\
          \midrule
          
          OAI CLIP~\cite{radford2021learning} & ViT-L/14
           &    85.7
           &    73.5
           &    98.7
           &    89.7
           &    95.4
           &    83.5
           &    96.2
           &    94.5
           &    94.4
           &    89.2
           &    5.7
           &    98.0
           &    62.0
           &    66.9
          \\
         \multirow{2}{*}{DFN-CLIP~\cite{fang2023data}} & VIT-L/14
          &    86.5
          &    75.5
          &    99.2
          &    93.2
          &    96.2
          &    85.8
          &    96.3
          &    96.4
          &    95.0
          &    89.8
          &    5.8
          &    98.3
          &    63.1
          &    66.8
          \\
          & ViT-H/14
          &    86.9
          &    76.4
          &    99.3
          &    93.9
          &    96.3
          &    87.0
          &    96.8
          &    96.7
          &    95.7
          &    90.5
          &    6.1
          &    98.8
          &    63.4
          &    68.1
           \\
        \multirow{2}{*}{SigLIP~\cite{zhai2023sigmoid}} & ViT-L/16
           &    86.5
           &    75.1
           &    98.5
           &    90.4
           &    96.1
           &    86.7
           &    96.7
           &    96.5
           &    93.1
           &    89.5
           &    4.5
           &    98.3
           &    61.7
           &    71.0
           \\
            & ViT-So400m/14
            &    87.3
            &    77.4
            &    98.8
            &    91.2
            &    96.5
            &    87.7
            &    96.7
            &    96.6
            &    93.3
            &    90.0
            &    4.6
            &    98.6
            &    64.4
            &    72.3
           \\
           \midrule
           & ViT-L/14
           &    86.6
           &    76.0
           &    99.1
           &    92.2
           &    95.7
           &    87.9
           &    96.3
           &    96.3
           &    93.7
           &    89.3
           &    5.6
           &    98.4
           &    60.7
           &    69.0
           \\
          & ViT-H/14
          &    87.5
          &    77.9
          &    99.3
          &    93.5
          &    96.3
          &    88.2
          &    96.6
          &    96.4
          &    93.3
          &    89.3
          &    5.8
          &    98.5
          &    62.2
          &    70.4
          \\
          & ViT-1B/14
          &    88.1
          &    79.7
          &    99.4
          &    94.1
          &    96.7
          &    88.4
          &    96.8
          &    96.5
          &    94.2
          &    89.0
          &    6.7
          &    98.8
          &    63.2
          &    71.7
          \\
          & ViT-3B/14
          &    88.5
          &    81.5
          &    99.5
          &    94.3
          &    96.8
          &    88.9
          &    97.1
          &    96.5
          &    93.5
          &    89.4
          &    7.3
          &    99.0
          &    64.2
          &    72.2
          \\
          \multirow{-5}{*}{\Ours} & ViT-3B/14$_{448\text{px}}$
          &    89.5
          & 85.9  & 99.5 & 94.5 & 97.4 & 89.0 & 97.4 & 96.7 & 93.4 & 89.9 & 9.5 & 98.9 & 66.1 & 74.8
          \\
    \end{tabular}} \caption{\textbf{Frozen trunk evaluation for recognition
    benchmarks.} We report the recognition performance of the \Ours family
    models when compared to a number of self-supervised and weakly-supervised
    state-of-the-art models. All models are evaluated using attentive probing
    with a frozen backbone. Unless otherwise specified, all \Ours models are trained at 224px resolution on 12B samples.}
    \label{tab:recognition}
    \vspace{-5mm}
\end{table*}

\subsection{AIMv2 vs.~Captioning}

We study the role of the image-level autoregressive objective in the
pre-training of \Ours. We compare the performance of models trained with the
multimodal autoregressive objective to ones trained only with language
supervision. The results are illustrated in \cref{fig:aimv2_vs_cap_plot}.

\cpar{Setup.} Unless specified otherwise, this investigation uses a ViT-H
backbone and 2 billion image-text pairs for pre-training. All models are trained
to convergence with a cosine learning rate schedule. We measure the IN-1k
top-1 accuracy after attentive probe with a frozen trunk.

\cpar{Results.} First, the image-level objective of \Ours consistently improves
performance compared to the captioning-only baseline. This is true even when
changing the model capacity and the size of the pre-training data. Moreover, we
see both approaches improving consistently when increasing the data size or
model capacity; however, we observe signs of plateauing for the captioning
baselines with scaling data which we do not observe with \Ours.

\section{Results}

\Ours is a generalist vision encoder that can be leveraged off-the-shelf for a
wide range of downstream tasks. We evaluate the performance of the \Ours family
across various tasks, including recognition, detection, captioning, and multiple
multimodal benchmarks.

\subsection{Image Recognition}

\cpar{Attentive probing.} We assess the quality of the \Ours models as
off-the-shelf backbones for recognition benchmarks which are outlined
in~\cref{tab:recognition_benchmarks}. To this end, we adopt the attentive
probing setting proposed by~\citet{yu2022coca}, where the vision encoder remains
frozen, and only an attentive probe classifier is trained on top of the last
layer features. The results are presented in~\cref{tab:recognition}. Detailed
hyperparameters used for the probing experiments are provided
in~\cref{tab:probe_hparams}. First, we observe that \Ours significantly
outperforms generative unsupervised methods such as MAE~\cite{he2021masked} and
AIM~\cite{el2024scalable}, even with much smaller capacity models. Compared to
DINOv2~\cite{oquab2023dinov2}, we find that both the similarly sized \Ours-1B
and the smaller \Ours-H provide competitive performance, outperforming DINOv2 on
several benchmarks including IN-1k, Food101, DTD, Cars, and with a particularly
large margin on Infographic. However, DINOv2 offers exceptional performance for
iNaturalist and fMoW. Furthermore, we find the performance of self-supervised
models on medical imaging benchmarks (\eg, RxRx1 and CAM17) noteworthy, as they
exhibit stronger performance compared to their weakly supervised counterparts.
This affirms the importance of self-supervised learning methods, particularly in
low-resource domains.

\ifarxiv

\begin{table}[tb]
    \centering
    \setlength{\tabcolsep}{10pt}
    \renewcommand{\arraystretch}{1}
    \resizebox{1\linewidth}{!}{
    \begin{tabular}{lccccccc}
        & \multicolumn{2}{c}{\textit{open-vocabulary}} &~& \multicolumn{3}{c}{\textit{referring expression}} \\
        \cline{2-3} \cline{5-7}
        model & COCO & LVIS && RefC & RefC+ & RefCg \\
        \shline
        OAI CLIP & 59.1 & 31.0 && 92.2 & 86.2 & 88.3     \\
        DFN-CLIP & 59.8 & 30.7 && 92.5 & 85.8 & 88.3     \\
        SigLIP &  58.8 & 30.5 && 92.3 & 86.1 & 88.4        \\
        DINOv2 & 60.1 & 30.8 && 92.2 & 85.9 & \textbf{89.1}        \\
        \Ours & \textbf{60.2} & \textbf{31.6} && \textbf{92.6} & \textbf{86.3} & 88.9        \\
     \end{tabular}} \caption{\textbf{Evaluation after finetuning on grounding
        dataset mixture.} We report the performance on mean average precision
        (AP) for open-vocabulary detection and Precision @1 for referring
        expression comprehension tasks.}
    \label{tab:grounding}
\end{table}

\begin{table}[tb]
    \vspace{-3mm}
    \centering
    \scalebox{0.97}{
        \tablestyle{2pt}{1.2}
        \begin{tabular}{lcc|cc|cc}
            model$~\rightarrow$ & Cap & \Ours & CapPa~\cite{tschannen2024image} & \Ours & \graytext{OAI CLIP} & \graytext{SigLIP} \\
            \shline
            Pre-train/LiT & 2B/3B & 2B/3B & 9B/3B & 12B/3B & \graytext{13B/-} & \graytext{40B/-} \\
            IN-1k top-1 & 75.0 & \textbf{75.3} & 76.4 & \textbf{77.0} & \graytext{75.5} & \graytext{80.4} \\
        \end{tabular}
    } \caption{\textbf{Zero-shot performance.} Comparison of different models
    with varying amounts of pre-training and LiT pairs, and their performance on
    IN1k. For CapPa, we compare to the number reported
    by~\citet{tschannen2024image}.}
    \label{tab:zero_shot}
\end{table}

    \begin{table}[tb]
    \vspace{-3mm}
    \centering
    \scalebox{1.1}{
        \tablestyle{8pt}{1}
        \begin{tabular}{lccc}
            test resolution $~\rightarrow$ & 224$\times$224  & 448$\times$448 & \textit{native} \\
            \shline
            \Ours-L$_{_\text{224px}}$ & \graytext{86.6}  & 84.8 & - \\
            \Ours-L$_{_\text{448px}}$ & 78.9 & \graytext{87.9} & - \\
            \Ours-L${_\text{native}}$ & 86.1 & 87.1 & 87.3 \\
        \end{tabular}
    } \caption{\textbf{\Ours with native apsect ratio and resolution.} We report
    the IN-1k top-1 performance of the native resolution \Ours-L model as
    compared to \Ours-L models that are pre-trained/finetuned for single
    dedicated resolution.}
    \label{tab:native_res}
\end{table}

\begin{table*}[h!]
    \centering
    \setlength{\tabcolsep}{5pt}
    \renewcommand{\arraystretch}{1}
    \resizebox{1\linewidth}{!}{
    \begin{tabular}{ll c ccccccccccccc}
        model & architecture & \# patches & \small{VQAv2} & \small{GQA} & \small{OKVQA}  & \small{TextVQA} & \small{DocVQA} & \small{InfoVQA} & \small{ChartQA} & \small{ScienceQA} & \small{COCO} & \small{TextCaps} & \small{NoCaps} & \small{SEED} & \small{MME$_\text{p}$}\\

        \shline

        OpenAI CLIP & ViT-L/14 & 576
        & 78.0
        & 72.0
        & 60.0
        & 47.5
        & 25.6
        & 21.8
        & 18.5
        & 73.8
        & 94.9
        & 75.3
        & 93.3
        & 70.1
        & 1481
         \\
        \multirow{2}{*}{SigLIP} & ViT-L/14 & 576
        & 76.9
        & 70.3
        & 59.3
        & 44.1
        & 16.9
        & 20.7
        & 14.4
        & 74.7
        & 93.0
        & 69.9
        & 92.7
        & 66.8
        & 1416
        \\
        & ViT-So400M/14 & 752
        & 77.7
        & 71.0
        & 60.1
        & 47.5
        & 19.2
        & 21.0
        & 14.7
        & 74.9
        & 94.6
        & 70.8
        & 94.5
        & 67.5
        & 1433
        \\
        DINOv2 & VIT-g/14 & 3034
        & 76.7
        & 72.7
        & 56.9
        & 15.1
        & 8.2
        & 19.7
        & 12.0
        & 69.5
        & 93.4
        & 42.1
        & 89.1
        & 68.9
        & 1423
         \\
         \midrule
        & ViT-L/14 & 576
        & 79.7
        & 72.5
        & 60.8
        & 53.6
        & 26.6
        & 22.8
        & 19.2
        & 74.1
        & 96.9
        & 81.1
        & 99.9
        & 71.8
        & 1472
        \\
        & ViT-H/14 & 576
        & 80.2
        & 72.8
        & 61.3
        & 55.5
        & 27.8
        & \textbf{23.1}
        & 19.9
        & 76.8
        & 99.6
        & 80.7
        & 102.7
        & 72.1
        & 1545
         \\
         & ViT-1B/14 & 576
        & 80.5
        & 73.0
        & 61.5
        & 56.8
        & 28.5
        & 22.1
        & 20.5
        & 76.4
        & 98.4
        & 82.6
        & 101.5
        & 72.7
        & 1508
        \\
        \multirow{-4}{*}{\Ours} & ViT-3B/14 & 576 & \textbf{80.9} &
        \textbf{73.3} & \textbf{61.7} & \textbf{58.2} & \textbf{30.4} & 23.0 &
        \textbf{22.6} & \textbf{77.3} & \textbf{100.3} & \textbf{83.8} &
        \textbf{102.9} & \textbf{72.9} & \textbf{1545} \end{tabular}}
        \caption{\textbf{Mutlimodal Evaluations.} We compare \Ours to
        state-of-the-art visual backbones for multimodal instruction tuning.
        Under comparable capacities, \Ours-L outperforms its counterparts on the
        majority of benchmarks. Additionally, scaling to the larger \Ours-3B
        model results in clear improvements, achieving the highest scores on
        nearly all benchmarks. All \Ours models use 336px resolution.}
    \label{tab:multimodal_results}
    \vspace{-5mm}
\end{table*}

\fi

Second, when compared to other vision-language pre-trained baselines, \Ours
exhibits highly competitive performance. For instance, at the ViT-Large
capacity, \Ours outperforms OAI CLIP on the majority of benchmarks and achieves
stronger performance than DFN-CLIP and SigLIP on several key benchmarks,
including IN-1k, iNaturalist, DTD, and Infographic. These results are
particularly impressive given that \Ours is trained using nearly a quarter of
the data required for training DFN-CLIP and SigLIP (12B \vs 40B), while also
being easier to train and scale. Finally, we find that scaling the capacity of
\Ours models consistently leads to a stronger performance with \Ours-3B
exhibiting the strongest result, in particular its variant finetuned for 448px
images which achieves 89.5\% top-1 accuracy on IN-1k with a frozen trunk.
Finally, in~\cref{fig:scaling_model_res} we observe a clear improvement to the
performance of IN-1k when scaling the model capacity and the image resolution in
conjunction. We provide more detailed results for the high-resolution finetuned
backbones  in~\cref{app:image_recognition}.

\ifarxiv
\else

\begin{table}[tb]
    \centering
    \setlength{\tabcolsep}{10pt}
    \renewcommand{\arraystretch}{1}
    \resizebox{1\linewidth}{!}{
    \begin{tabular}{lccccccc}
        & \multicolumn{2}{c}{\textit{open-vocabulary}} &~& \multicolumn{3}{c}{\textit{referring expression}} \\
        \cline{2-3} \cline{5-7}
        model & COCO & LVIS && RefC & RefC+ & RefCg \\
        \shline
        OAI CLIP & 59.1 & 31.0 && 92.2 & 86.2 & 88.3     \\
        DFN-CLIP & 59.8 & 30.7 && 92.5 & 85.8 & 88.3     \\
        SigLIP &  58.8 & 30.5 && 92.3 & 86.1 & 88.4        \\
        DINOv2 & 60.1 & 30.8 && 92.2 & 85.9 & \textbf{89.1}        \\
        \Ours & \textbf{60.2} & \textbf{31.6} && \textbf{92.6} & \textbf{86.3} & 88.9        \\
     \end{tabular}} \caption{\textbf{Evaluation after finetuning on grounding
        dataset mixture.} We report the performance on mean average precision
        (AP) for open-vocabulary detection and Precision @1 for referring
        expression comprehension tasks.}
    \label{tab:grounding}
\end{table}

\begin{table}[tb]
    \vspace{-3mm}
    \centering
    \scalebox{0.97}{
        \tablestyle{2pt}{1.2}
        \begin{tabular}{lcc|cc|cc}
            model$~\rightarrow$ & Cap & \Ours & CapPa~\cite{tschannen2024image} & \Ours & \graytext{OAI CLIP} & \graytext{SigLIP} \\
            \shline
            Pre-train/LiT & 2B/3B & 2B/3B & 9B/3B & 12B/3B & \graytext{13B/-} & \graytext{40B/-} \\
            IN-1k top-1 & 75.0 & \textbf{75.3} & 76.4 & \textbf{77.0} & \graytext{75.5} & \graytext{80.4} \\
        \end{tabular}
    } \caption{\textbf{Zero-shot performance.} Comparison of different models
    with varying amounts of pre-training and LiT pairs, and their performance on
    IN1k. For CapPa, we compare to the number reported
    by~\citet{tschannen2024image}.}
    \label{tab:zero_shot}
\end{table}

    \begin{table}[tb]
    \vspace{-3mm}
    \centering
    \scalebox{1.1}{
        \tablestyle{8pt}{1}
        \begin{tabular}{lccc}
            test resolution $~\rightarrow$ & 224$\times$224  & 448$\times$448 & \textit{native} \\
            \shline
            \Ours-L$_{_\text{224px}}$ & \graytext{86.6}  & 84.8 & - \\
            \Ours-L$_{_\text{448px}}$ & 78.9 & \graytext{87.9} & - \\
            \Ours-L${_\text{native}}$ & 86.1 & 87.1 & 87.3 \\
        \end{tabular}
    } \caption{\textbf{\Ours with native apsect ratio and resolution.} We report
    the IN-1k top-1 performance of the native resolution \Ours-L model as
    compared to \Ours-L models that are pre-trained/finetuned for single
    dedicated resolution.}
    \label{tab:native_res}
\end{table}

\fi

\begin{table}[t!]
    \centering
    \scalebox{1}{
    \tablestyle{8pt}{1}
    \begin{tabular}{llccc}
         model & architecture &  0-shot & 4-shot & 8-shot \\
        \midrule
        OAI CLIP~\cite{mckinzie2024mm1} & ViT-L/14 &39.3 & 62.2 & 66.1  \\
        DFN-CLIP~\cite{mckinzie2024mm1} & ViT-H/14 & \bf 40.9 & 62.5 & 66.4  \\
        \Ours & ViT-L/14 & 39.6 & \bf 63.8 & \bf 67.2  \\
    \end{tabular}} \caption{\textbf{ICL few-shot performance.} We report the
    in-context few-shot performance averaged across a wide range of benchmarks
    as detailed in~\cref{sec:multimodal_results}. The results for DFN-CLIP and
    OAI CLIP are as reported by~\citet{mckinzie2024mm1}.\label{tab:icl}}
\end{table}

\cpar{Zero-shot via LiT Tuning.} We investigate the compatibility of \Ours
backbones with LiT~\cite{zhai2022lit}, extending its application to zero-shot
settings. We report the IN-1k zero-shot performance in~\cref{tab:zero_shot}.
First, we observe that \Ours, with the multimodal autoregressive objective,
shows a modest improvement compared to the captioning-only baseline, even in
this setting. Furthermore, an \Ours-L model trained for a longer duration
exhibits favorable performance compared to the results reported
by~\citet{tschannen2024image} for CapPa. Overall, our model demonstrates strong
zero-shot performance, outperforming OAI CLIP~\cite{radford2021learning}, yet
still lagging behind dedicated models like SigLIP that are trained for a longer
schedule with 40B image-text pairs.

\cpar{Native resolution.}
We finetune \Ours to process images with a wide range of resolutions and aspect
ratios as detailed in~\cref{sec:post_training}. In order to assess the quality
of this stage of post-training, we compare the performance of an \Ours encoder
adapted for native resolution to models that are tuned for one specific
resolution in~\cref{tab:native_res}. We observe that \Ours-L$_{\text{native}}$
provides a strong performance across a wide range of resolutions off-the-shelf,
experiencing only a minor degradation in performance to the dedicated models.
Additionally, evaluating our model using the original native resolutions of the
IN-1k validation set images  yields a robust accuracy of 87.3\%, confirming that
\Ours maintains exceptional recognition performance while offering high
flexibility in both aspect ratio and resolution.

\definecolor{OAI_CLIP}{HTML}{879838}
\definecolor{SigLIP}{HTML}{DDA263}
\definecolor{AIMv2}{HTML}{004B95}

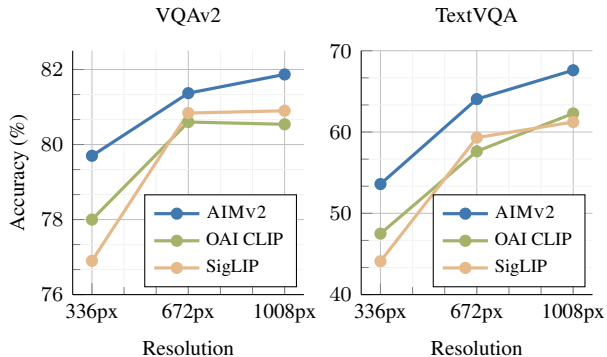
\begin{figure}[t!]
    \centering
    \captionsetup{type=figure}
    \begin{subfigure}[t]{0.49\linewidth}
        \begin{tikzpicture}
            \begin{axis}[
                title={VQAv2},
                title style={font=\footnotesize},
                legend pos=south east,
                grid=both,
                grid style={line width=.1pt, draw=gray!10},
                major grid style={line width=.2pt,draw=gray!50},
                minor tick num=2,
                axis x line*=bottom,
                axis y line*=left,
                xtick={1, 2, 3},
                xticklabels={336px, 672px, 1008px},
                ymin=76, ymax=82.5,
                xlabel={Resolution},
                ylabel={Accuracy (\%)},
                ylabel style={align=center, font=\footnotesize, yshift=-0.5ex},
                xlabel style={font=\footnotesize},
                yticklabel style={font=\footnotesize},
                xticklabel style={font=\footnotesize},
                legend style={cells={align=left}, font=\scriptsize},
                legend cell align={left},
                mark options={solid},
                width=1.14\linewidth,
                height=1.9in,
            ]

            \addplot[mark=oplus*, AIMv2!75, line width=1.2pt, mark size=1.7pt] coordinates {
                (1, 79.7)
                (2, 81.37)
                (3, 81.869)
            };
            \addlegendentry{\Ours}

            \addplot[mark=oplus*, OAI_CLIP!75, line width=1.2pt, mark size=1.7pt] coordinates {
                (1, 78.0)
                (2, 80.60)
                (3, 80.54)
            };
            \addlegendentry{OAI CLIP}

            \addplot[mark=oplus*, SigLIP!75, line width=1.2pt, mark size=1.7pt] coordinates {
                (1, 76.9)
                (2, 80.84)
                (3, 80.90)
            };
            \addlegendentry{SigLIP}

            \end{axis}
        \end{tikzpicture}
    \end{subfigure}
    \hfill
    \begin{subfigure}[t]{0.49\linewidth}
        \begin{tikzpicture}
            \begin{axis}[
                title={TextVQA},
                title style={font=\footnotesize},
                legend pos=south east,
                grid=both,
                grid style={line width=.1pt, draw=gray!10},
                major grid style={line width=.2pt,draw=gray!50},
                minor tick num=2,
                axis x line*=bottom,
                axis y line*=left,
                xtick={1, 2, 3},
                xticklabels={336px, 672px, 1008px},
                ymin=40, ymax=70,
                xlabel={Resolution},
                ylabel style={align=center, font=\footnotesize, yshift=-0.5ex},
                xlabel style={font=\footnotesize},
                yticklabel style={font=\footnotesize},
                xticklabel style={font=\footnotesize},
                legend style={cells={align=left}, font=\scriptsize},
                legend cell align={left},
                mark options={solid},
                width=1.14\linewidth,
                height=1.9in,
            ]

            \addplot[mark=oplus*, AIMv2!75, line width=1.2pt, mark size=1.7pt] coordinates {
                (1, 53.6)
                (2, 64.05)
                (3, 67.60)
            };
            \addlegendentry{\Ours}

            \addplot[mark=oplus*, OAI_CLIP!75, line width=1.2pt, mark size=1.7pt] coordinates {
                (1, 47.5)
                (2, 57.63)
                (3, 62.29)
            };
            \addlegendentry{OAI CLIP}

            \addplot[mark=oplus*, SigLIP!75, line width=1.2pt, mark size=1.7pt] coordinates {
                (1, 44.1)
                (2, 59.31)
                (3, 61.23)
            };
            \addlegendentry{SigLIP}

            \end{axis}
        \end{tikzpicture}
    \end{subfigure}       
    \caption{\textbf{Impact of Scaling Resolution.} The performance boost
    achieved by\Ours persists after scaling input resolution via
    tiling~\citet{liu2024improvedllava, lin2023sphinx} compared to popular
    vision backbones for VLMs such as OAI CLIP and SigLIP.}
    \label{fig:scaling_res_mm}
\end{figure}
\begin{figure*}[htb]
    \hspace{-5mm}\subfloat[\textbf{\scriptsize{Vicuna 1.5 + Llava mixture}}\vspace{-4mm}]{
    \definecolor{CustomA}{HTML}{879838}
\definecolor{CustomB}{HTML}{DDA263}
\definecolor{CustomC}{HTML}{004B95}

\begin{tikzpicture}
    \begin{axis}[
        ybar,
        bar width=8pt,
        width=\textwidth,
        height=1in,
        width=1.1\linewidth,
        ymin=0,
        symbolic x coords={VQAv2, GQA, OKVQA, TextVQA, DocVQA, InfoVQA, ChartQA, ScienceQA, COCO, TextCaps, NoCaps, MMEp},
        yticklabel style={font=\scriptsize},
        xticklabel style={font=\scriptsize},
        ylabel style={align=center, font=\scriptsize, yshift=-0.5ex},
        xlabel style={font=\scriptsize},
        nodes near coords style={font=\tiny, text=black},
        cycle list={{CustomA!50},{CustomB},{CustomC}},
        xtick=data,
        nodes near coords,
        legend style={at={(0.5,1.7)}, anchor=north,legend columns=-1, font=\scriptsize},
        enlarge x limits={abs=1cm},
        axis line style={draw=none},
        tick style={draw=none}, 
        ytick=\empty, 
    ]

    \addplot+[fill=CustomA!50] coordinates {
        (VQAv2, 77)
        (GQA, 70)
        (OKVQA, 57)
        (TextVQA, 46)
        (DocVQA, 25)
        (InfoVQA, 23)
        (ChartQA, 18)
        (ScienceQA, 69)
        (COCO,93)
        (TextCaps, 74)
        (NoCaps, 91)
        (MMEp, 70)
    };
    \addplot+[fill=CustomB] coordinates {
        (VQAv2, 76)
        (GQA, 69)
        (OKVQA, 57)
        (TextVQA, 48)
        (DocVQA, 23)
        (InfoVQA, 22)
        (ChartQA, 15)
        (ScienceQA, 69)
        (COCO,93)
        (TextCaps, 72)
        (NoCaps, 91)
        (MMEp, 68)
    };

    \addplot+[fill=CustomC!90] coordinates {
        (VQAv2, 79)
        (GQA, 71)
        (OKVQA, 58)
        (TextVQA, 53)
        (DocVQA, 28)
        (InfoVQA, 23)
        (ChartQA, 21)
        (ScienceQA, 69)
        (COCO,97)
        (TextCaps, 79)
        (NoCaps, 100)
        (MMEp, 72)
    };
    \legend{OAI CLIP, SigLIP, AIMV2}
    \end{axis}
\end{tikzpicture}
    }

    \hspace{-5mm}\subfloat[\textbf{\scriptsize{Llama 3.0 + Cambrian 7M}}]{
    \definecolor{CustomA}{HTML}{879838}
\definecolor{CustomB}{HTML}{DDA263}
\definecolor{CustomC}{HTML}{004B95}

\begin{tikzpicture}
    \begin{axis}[
        ybar,
        bar width=8pt,
        width=\textwidth,
        height=1in,
        width=1.1\linewidth,
        ymin=0,
        symbolic x coords={VQAv2, GQA, OKVQA, TextVQA, DocVQA, InfoVQA, ChartQA, ScienceQA, COCO, TextCaps, NoCaps, MMEp},
        yticklabel style={font=\scriptsize},
        xticklabel style={font=\scriptsize},
        ylabel style={align=center, font=\scriptsize, yshift=-0.5ex},
        xlabel style={font=\scriptsize},
        nodes near coords style={font=\tiny, text=black},
        cycle list={{CustomA!50},{CustomB},{CustomC}},
        xtick=data,
        nodes near coords,
        legend style={at={(0.5,1.7)}, anchor=north,legend columns=-1, font=\scriptsize},
        enlarge x limits={abs=1cm},
        axis line style={draw=none},
        tick style={draw=none}, 
        ytick=\empty, 
    ]

    \addplot+[fill=CustomA!50] coordinates {
        (VQAv2, 74)
        (GQA, 71)
        (OKVQA, 60)
        (TextVQA, 53)
        (DocVQA, 51)
        (InfoVQA, 34)
        (ChartQA,51)
        (ScienceQA, 80)
        (COCO,95)
        (TextCaps, 78)
        (NoCaps, 96)
        (MMEp, 80)
    };
    \addplot+[fill=CustomB] coordinates {
        (VQAv2, 75)
        (GQA, 71)
        (OKVQA, 61)
        (TextVQA, 56)
        (DocVQA, 51)
        (InfoVQA,34)
        (ChartQA, 48)
        (ScienceQA, 80)
        (COCO,94)
        (TextCaps, 80)
        (NoCaps, 96)
        (MMEp, 78)
    };

    \addplot+[fill=CustomC!90] coordinates {
        (VQAv2, 76)
        (GQA, 72)
        (OKVQA, 61)
        (TextVQA, 58)
        (DocVQA, 50)
        (InfoVQA, 35)
        (ChartQA, 52)
        (ScienceQA, 79)
        (COCO, 96)
        (TextCaps, 82)
        (NoCaps, 98)
        (MMEp, 80)
    };
    \end{axis}
\end{tikzpicture}
    }

    \caption{\textbf{Instruction tuning under different settings.} We evaluate
    instruction-tuned models across combinations of LLM decoders and tuning data
    mixtures. In all settings, \Ours consistently outperforms or matches SigLIP
    and OAI CLIP on most benchmarks. All models use a ViT-L backbone with 336px
    images. For better readability, we present normalized MME$_p$ scores by
    dividing the raw scores by 2000.}
    \label{fig:mm_barcharts}
    \vspace{-5mm}
\end{figure*}
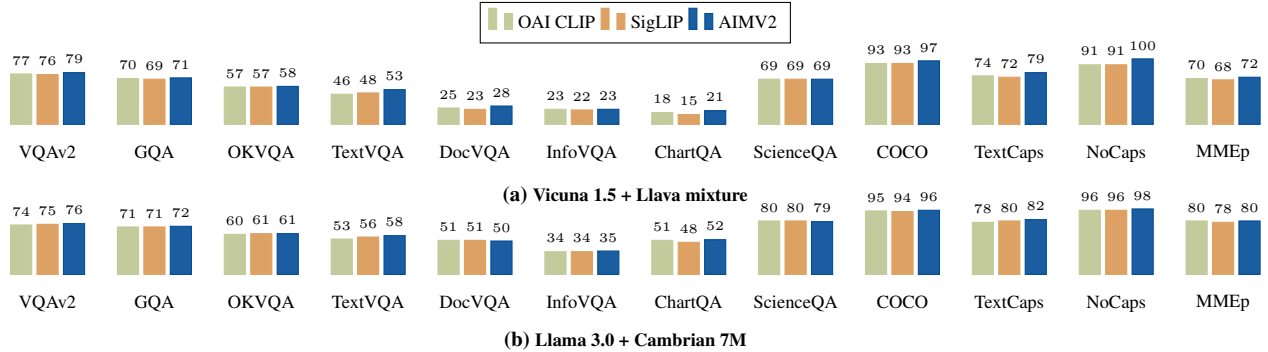

\subsection{Object Detection and Grounding}

\noindent To further demonstrate the capabilities of \Ours, we evaluate its
performance on tasks such as Open-Vocabulary Detection (OVD) and Referring
Expression Comprehension (REC). We follow the model architecture introduced by
MM-Grounding-DINO~\cite{liu2024groundingdinomarryingdino,mmgdino2024} but adapt
ViT-L through the ViTDet \cite{li2022exploring} formulation as the vision
backbone. Our results are presented in~\cref{tab:grounding}. For OVD
capabilities we evaluate on COCO~\cite{coco2014} and
LVIS~\cite{gupta2019lvisdatasetlargevocabulary},
while for REC, we evaluate on RefCOCO (RefC)~\cite{kazemzadeh2014referitgame},
RefCOCO+ (RefC+)~\cite{yu2016modelingcontextreferringexpressions}, and RefCOCOg
(RefCg)~\cite{mao2016generation}. All models were trained on the following
datasets: Objects365v1~\cite{o365v1}, Flickr-30k
Entities~\cite{flickr30k,flickr30kentities}, GQA~\cite{gqa2019}, COCO17
~\cite{coco2014}, and
RefCOCO~\cite{kazemzadeh2014referitgame,yu2016modelingcontextreferringexpressions,mao2016generation}.
During DINOv2 training we fix the window size to 16
~\cite{li2022exploringplainvisiontransformer} to ensure fixed compute cost
across backbones. Our results indicate that \Ours outperforms DINOv2 as well as
other vision-language pre-trained models on all benchmarks but one, showing
particularly strong performance on LVIS. We present additional localization and
grounding results including closed-vocabulary detection and instance
segmentation as well as ablations on varying window sizes
in~\cref{app:grounding}.

\subsection{Multimodal Understanding}

Vision encoders play a crucial role in advancing large multimodal
models~\cite{liu2024improvedllava,zhu2024minigpt4,bai2023qwenvl,mckinzie2024mm1,tong2024cambrian}.
To quantify the performance of \Ours models in this setting, we perform a
multimodal instruction tuning stage similar to~\citet{liu2024improvedllava}.
Additionally, we explore the few-shot In-Context Learning (ICL) setting after
large-scale multimodal pretraining similar to~\citet{mckinzie2024mm1}.

\subsubsection{Multimodal Instruction Tuning}

\cpar{Setup.} We place a 2-layer MLP connector between the vision encoder
(\eg,~\Ours-L) and the LLM (\eg,~Llama~3.0). The parameters of the vision
encoder are frozen during this stage. Contrary to~\citet{liu2024improvedllava},
we train the connector and the LLM jointly in a single stage. However, we scale
up the learning rate for the connector by a factor of 8. We found this strategy
to be simpler while leading to comparable results. We detail the evaluation
datasets, task prompts, and hyperparameters used during this stage
in~\cref{app:sft}. Unless mentioned otherwise, we use the Llava SFT
mixture~\cite{liu2024improvedllava} and Llama-3.0 8B LLM decoder~\cite{llama3}. We train all
models for a single epoch. 

\cpar{Evaluation.} We evaluate the instruction-tuned models across various
question answering benchmarks covering general knowledge, text-rich images,
scientific domains, and captioning. The results for \Ours and several
baselines are presented in~\cref{tab:multimodal_results}.
Notably, our smallest model, \Ours-L, outperforms OAI CLIP, SigLIP, and DINOv2
on most benchmarks, even when the baselines use larger capacities or
higher input resolutions. Furthermore, performance consistently improves with
increasing the \Ours backbone capacity, with the \Ours-3B model achieving the best
performance across all benchmarks except one.

\cpar{Varying the LLM and Data Mixture.} In addition to the canonical setting
reported in~\cref{tab:multimodal_results}, we evaluate whether \Ours can provide
similar gains compared to popular vision encoders across various combinations of
LLM decoders and instruction tuning data mixtures. Specifically, we perform the
instruction tuning stage under the following settings: (1) Llama 3.0 with the
Cambrian data mixture~\cite{tong2024cambrian} and (2) Vicuna 1.5~\cite{vicuna}
with the Llava SFT mixture. We present the results for \Ours-L alongside
similarly sized OAI CLIP and SigLIP backbones in~\cref{fig:mm_barcharts}. Across
all settings, \Ours provides a stronger, or at worst on par, performance
compared the OAI CLIP and SigLIP. These findings further affirm the robustness
and compatibility of \Ours within diverse multimodal pipelines.

\cpar{High-Resolution via Tiling.} One of the most popular strategies to enhance
the performance of vision-language models is increasing the image resolution.
This can be achieved through a tiling
strategy~\cite{lin2023sphinx,liu2024improvedllava,shi2024we}, where a
high-resolution image is divided into a number of equally sized crops that match
the pre-training resolution of the available backbones (\eg, 224px or 336px). We
investigate the compatibility of \Ours with this strategy. Specifically, we use
a crop size of 336px and evaluate our pipeline on 672px and 1008px images
corresponding to 2$\times$2 and 3$\times$3 grids respectively. The results for
\Ours, SigLIP, and OAI CLIP are presented in~\cref{fig:scaling_res_mm}. We
observe that the performance of all methods improves with higher resolutions,
with a significant improvement for TextVQA. Notably, \Ours maintains its
advantage over the baselines in high-resolution tiling settings, demonstrating
its versatility.

\subsubsection{Multimodal In-Context Learning}
\label{sec:multimodal_results} We also evaluate \Ours in a large-scale
multimodal pre-training setting. Following the pre-training recipe as
MM1~\cite{mckinzie2024mm1}, we simply replace the vision encoder with \Ours.
Given that this model is pre-trained using interleaved image-text documents, it
enables in-context evaluations~\cite{alayrac2022flamingo}. We report
the ICL performance in~\cref{tab:icl}. Specifically, we report the average
0-shot, 4-shot, and 8-shot performance across the following benchmarks:
COCO~\cite{chen2015microsoft}, NoCaps~\cite{agrawal2019nocaps},
TextCaps~\cite{sidorov2020textcaps}, VQAv2~\cite{goyal2017making},
TextVQA~\cite{singh2019towards}, VizWiz~\cite{gurari2018vizwiz},
GQA~\cite{hudson2019gqa}, and OK-VQA~\cite{marino2019ok}. Our results
demonstrate that \Ours achieves the best performance in the 4-shot and 8-shot
settings, surpassing the higher capacity DFN-CLIP adopted by the MM1 series.
This highlights the compatibility and effectiveness of \Ours in leveraging ICL
in a large-scale multimodal setup.

\begin{table*}[h!]
        \centering
         \scalebox{0.9}{ \subfloat[ \textbf{Objective}.
                \label{tab:aimv1_aimv2_cap_ablation}
                ]{
                    \centering
                        \begin{minipage}[t]{0.43\linewidth}{
                    \begin{center}
                    \tablestyle{5pt}{1.0}
                    \begin{tabular}{l  cc ccc}
                        model & pre-train attn. &       IN-1k &            VQAv2
                        &       TextVQA  \\
                         \shline
                                          \aim & \textit{prefix}  &
                                          72.0 &    65.4 &  12.7      \\
                          \multirow{2}{*}{Cap} & \textit{bidir}   &
                          85.1 &    76.2 &  34.4      \\
                                               & \textit{prefix}  &
                                               85.4 &    76.8 &  36.5    \\
                               \grayrow  \Ours & \textit{prefix}  &
                               \textbf{85.6} &   \textbf{76.9}  &  \textbf{37.5}
                               \\
                     \end{tabular}
                        \end{center}}\end{minipage} } \hfill \subfloat[
                \textbf{\Ours \vs CLIP.}.
                        \label{tab:aimv2_vs_clip}
                        ]{
                                \centering
                                \begin{minipage}[t]{0.32\linewidth}{\begin{center}
                                \tablestyle{5pt}{1.0}
                                \begin{tabular}{l  cccc}
                                        model & bsz &    IN-1k &
                                        VQAv2 &       TextVQA  \\
                                        \shline
                                        \multirow{2}{*}{CLIP} & 8k & 84.6 & 74.1
                                        & 24.6 \\
                                        & 16k & 85.2 & 74.8 & 26.3\\
                                        CapPa & 8k & 84.7 & 75.1 & 30.6 \\
                                        \grayrow \Ours          & 8k &
                                        \textbf{85.6} &   \textbf{76.9}  &
                                        \textbf{37.5} \\
                                \end{tabular}
                \end{center}}\end{minipage} } \subfloat[ \textbf{Criteria
                Weights}.
                \label{tab:criteria_weights}
                ]{
                        \vspace{2.2mm}
                        \centering
                        \begin{minipage}[t]{0.3\linewidth}{\begin{center}
                        \tablestyle{7pt}{1.1}
                        \begin{tabular}{l  ccc}
                             $\alpha$ &  IN-1k &            VQAv2 &
                             TextVQA  \\
                            \shline
                             0.2           & \textbf{85.6}  & 76.7 & 37.4 \\
                             \grayrow  0.4 &  \textbf{85.6} &   \textbf{76.9}  &
                             \textbf{37.5} \\
                             0.6           & \textbf{85.6}  & 76.7 & 37.4 \\
                        \end{tabular}
                        \end{center}}\end{minipage} } }

        \scalebox{0.95}{ \subfloat[ \textbf{Decoder Architecture}.
                \label{tab:decoder_arch}
                ]{
                        \vspace{2.8mm}
                        \centering
                        \begin{minipage}[t]{0.35\linewidth}{
                        \begin{center}
                            \tablestyle{7pt}{1.1}                
                            \begin{tabular}{l  ccc}
                                                            &  IN-1k &
                                                            VQAv2 &
                                                            TextVQA  \\
                                \shline
                                \textit{separate}          &
                                \textbf{85.6} & \textbf{77.1} & 37.2\\
                         \grayrow \textit{joint}            &  \textbf{85.6} &
                         76.9  &  \textbf{37.5} \\
                            \end{tabular}
                               \end{center}}
                  \end{minipage}
          } \subfloat[ \textbf{Decoder Width}.
            \label{tab:decoder_width}
            ]{
                \centering
                \begin{minipage}{0.33\linewidth}{\begin{center}
                        \tablestyle{7pt}{1}
                        \begin{tabular}{l ccc}
                            width & IN-1k &            VQAv2 &       TextVQA  \\
                            \shline
                            512  &   85.3 & 76.2 & 35.9 \\ 
                   \grayrow 1024 &  \textbf{85.6} &   \textbf{76.9}  &
                   \textbf{37.5} \\
                            1536 &   85.1 &  \textbf{76.9} & 36.9  \\
                        \end{tabular}
                \end{center}}\end{minipage} } \subfloat[ \textbf{Decoder Depth}.
            \label{tab:decoder_depth}
            ]{
                \centering
                \begin{minipage}{0.3\linewidth}{\begin{center}
                        \tablestyle{7pt}{1}
                        \begin{tabular}{l ccc}
                            depth &  IN-1k &            VQAv2 &       TextVQA
                            \\
                            \shline
                            8  &          85.5 & 76.7 & 37.0 \\ 
                   \grayrow 12 &  \textbf{85.6} &   \textbf{76.9}  &
                   \textbf{37.5} \\
                            16 & \textbf{85.6} &  \textbf{76.9} & 36.6  \\
                        \end{tabular}
                \end{center}}\end{minipage} } }
        
            \caption{\textbf{Ablations.} We ablate a number of design choices
            for \Ours and how they impact performance on key recognition and
            multimodal benchmarks. This includes \textbf{(a)} the contribution
            of the visual and textual objectives, \textbf{(b)} comparison to
            other popular objectives,  \textbf{(c)} the optimal balancing
            between the losses, \textbf{(d-f)} the architecture of the
            multimodal decoder. All models are trained at 224px resolution.}    
            \label{tab:ablations}
            \vspace{-5mm}
         \end{table*}

\section{Ablation Study}

In this section, we investigate various design choices and present the
trade-offs associated with each. The results of our study are summarized
in~\cref{tab:ablations}.

\cpar{Setup.} The default setting for this ablation study utilizes a ViT-Large
vision encoder and 2 billion image-text pairs during pre-training. We measure
the IN-1k top-1 accuracy after attentive probing, as well as the
question-answering accuracy on the validation sets of
VQAv2~\cite{goyal2017making} and TextVQA~\cite{singh2019textvqa} following
instruction tuning, as described in~\cref{sec:post_training}. All experiments
reported in this ablation study employ images with 224 $\times$ 224 resolution.
The metrics selected for this study provide a comprehensive view of the models'
capabilities, encompassing recognition, general question answering, and
text-rich question answering.

\cpar{Pre-training Objective.} The pre-training objective of \Ours comprises a
combination of image-level and text-level autoregressive objectives. We evaluate
the performance of each objective independently and assess the impact of
combining them, as presented in~\cref{tab:aimv1_aimv2_cap_ablation}. First, we
observe that utilizing only the image-level objective (\ie, AIMv1) results in
weaker performance compared to models that incorporate the captioning objective.
This is expected, given that AIMv1 operates in an unsupervised manner and
demands higher-capacity models to achieve optimal performance, as highlighted
by~\citet{el2024scalable}. Second, for the captioning-only model, using prefix
attention within the vision encoder yields superior performance compared to
fully bidirectional attention. We hypothesize that prefix attention facilitates
the encoding of maximally informative contexts even from partial images, as such
contexts are utilized by subsequent visual and textual tokens. However, this
hypothesis warrants further investigation, which is beyond the scope of this
work and is reserved for future research. Finally, we find that combining the
image-level and text-level objectives in \Ours leads to an improved performance,
particularly noticeable for TextVQA.

\cpar{\Ours \vs CLIP \vs CapPa.} In~\cref{tab:aimv2_vs_clip}, we evaluate the
performance of models trained with the \Ours objective in comparison to other
popular vision-language pre-training objectives, specifically
CLIP~\cite{radford2021learning} and CapPa~\cite{tschannen2024image}. All models
are trained using identical architectures, incorporating SwiGLU and RMSNorm, and
are pre-trained using the same dataset of image-text pairs. Notably, since CLIP
pre-training benefits from larger batch sizes, we report CLIP results using both
8k and 16k batch sizes to ensure a fair comparison. Our findings indicate that,
under comparable settings, \Ours consistently outperforms both CLIP and CapPA by
a significant margin, particularly on the TextVQA benchmark. This performance is
especially noteworthy given the simplicity and scalability \Ours.

\cpar{Multi-task balancing.} We examine whether the pre-training of \Ours is
sensitive to the balancing between the image-level and text-level objectives
in~\cref{tab:criteria_weights}. We vary the hyperaparmeter $\alpha$, as
described in~\cref{sec:pretraining_method}, and we observe only minor
fluctuations in performance around the optimal value of 0.4 across the three
benchmarks.

\cpar{Joint vs. Separate Decoders.} In the \Ours architecture, we opt for a
multimodal joint decoder instead of employing dedicated decoders for each
modality. In~\cref{tab:decoder_arch}, we examine the performance of an \Ours
variant that utilizes two dedicated decoders. Using a single joint decoder
achieves comparable results to using separate decoders while offering
greater simplicity and efficiency during pre-training. This advantage proves
valuable when scaling data and model capacity.

\cpar{Decoder architecture.} Finally, we examine the capacity of the multimodal
decoder, as detailed in~\cref{tab:decoder_width} and~\cref{tab:decoder_depth}.
We find that performance is more sensitive to changes in decoder capacity when
scaling the width compared to scaling the depth. Additionally, we observe that
increasing the decoder capacity beyond a certain threshold, whether by scaling
width or depth, leads to a decline in performance. This observation is
consistent with the findings of~\citet{tschannen2024image} for captioning-only
models.

\ifarxiv
    \section{Related Works}
\label{sec:related_works}

\cpar{Autoregressive pre-training}. Autoregressive modeling has been a
foundational idea in machine learning and statistics for decades, long before
deep learning~\cite{box2015time}. However, it has been popularized and scaled by
works such as
GPT~\cite{radford2018improving,radford2019language,brown2020language}, and
LLaMAs~\cite{touvron2023llama,touvron2023llama2,dubey2024llama} which have
demonstrated the power of autoregressive pre-training in natural language
processing tasks. In vision, autoregressive principles have been applied through
models like iGPT~\cite{pmlr-v119-chen20s}, which flattens images into a sequence
of discretized pixels and then treats them analogously to language tokens.
Similarly, \citet{Yu2021VectorquantizedIM} also discretize the patches with a
VQGAN model~\cite{esser2021taming} and then predicts them autoregressively.
AIM~\cite{el2024scalable} brings back the more practical continuous approach and
scales to very large vision models. However, AIM still lags behind other state
of the art models in sheer performance, as it uses vision-only data and requires
large model capacities to perform optimally. This paper addresses these
limitations by introducing multimodal pre-training in the \Ours family.
Concurrent works
\cite{yu2023scaling,team2024chameleon, wu2024vila, xie2024show,lu2024unified,wang2022ofa,shukor2023unival}
have also investigated similar multimodal autoregressive approaches that predict
text and images. However, they often focus on multimodal generation quality rather
than representation quality, and therefore use discrete tokens or leverage diffusion models~\cite{rombach2022high} as
decoders~\cite{li2024mini, sun2024generative,
sun2023generative}. 

\cpar{Pre-training in vision}. For many years, the computer vision community
predominantly focused on supervised pre-training~\cite{kolesnikov2020big,
sun2017revisiting, ridnik2021imagenet}, with
ImageNet~\cite{krizhevsky2012imagenet} checkpoints serving as the backbone for
most visual tasks. This eventually hit a wall in terms of scalability, as labels
are expensive to acquire. The community therefore focused on self-supervised
methods. Earlier models used pre-text tasks such as rotation prediction and
patch deshuffling~\cite{gidaris2018unsupervised, noroozi2016unsupervised,
zhang2016colorful}. More sophisticated models like SimCLR~\cite{chen2020simple},
BYOL~\cite{grill2020bootstrap}, SwAV~\cite{caron2020unsupervised} and
DINO~\cite{caron2021emerging} leverage variations of contrastive learning to
train models that are quasi-invariant to a broad range of image augmentations.
This turns out to learn strong and general visual representations without
supervision. However they require carefully handcrafted data augmentations,
which also makes them computationally expensive, especially at scale. On the
other hand, MAE and BEiT~\cite{he2021masked, bao2021beit} introduced masking
strategies to reconstruct input data, reducing the reliance on augmentations and
increasing efficiency but sacrificing performance. In practice, the best
performing self-supervised vision-only models use a mixture of augmentations and
masking~\cite{oquab2023dinov2,assran2023self,zhou2021ibot}. Unfortunately, they
are challenging to scale as they still need multiple forward passes for each
training step. AIM~\cite{el2024scalable} departs from these methods by employing
a reconstruction-based autoregressive framework that exhibits strong scalability
but requires high capacity models to attain optimal performance. Leveraging
large-scale, noisy, weakly supervised datasets from the
internet~\cite{fang2023data,schuhmann2021laion,byeon2022coyo}, an efficient
paradigm emerged that aligns vision and text features through contrastive
learning~\cite{radford2021learning,jia2021scaling}. Nevertheless, CLIP-like
models require large batch sizes and meticulous dataset
filtering~\cite{fang2023data,schuhmann2021laion}. Subsequent works, such as
SigLIP~\cite{zhai2023sigmoid}, EVA CLIP~\cite{sun2023eva}, and
\citet{fini2023improved}, have addressed these issues by optimizing training
processes and improving data filtering~\cite{fang2023data}. Unlike these
approaches, \Ours does not perform explicit feature space alignment but aligns
training objectives through autoregressive modeling for better multimodal
synergy.

\cpar{Captioning.} Image captioning has been extensively studied prior to the
computer vision literature. Early
works~\cite{karpathy2015deep,vinyals2015show,xu2015show} focused on aligning
visual features with text to generate descriptions using CNNs and RNNs.
VirTex~\cite{desai2021virtex} and ICMLM~\cite{sariyildiz2020learning} utilize
captioning for visual pre-training. SimVLM~\cite{wang2021simvlm} employs a
PrefixLM approach, encoding images and partial text tokens with a multimodal
encoder and decoding the remaining text. LEMON~\cite{hu2022scaling} scales the
language model in both parameters and dataset size. Approaches such as
~\cite{li2022blip,li2021align} combines generative captioning with
discriminative contrastive objectives to enhance multimodal learning, which led
to scaling to billion-parameter models
~\cite{yu2022coca,kuo2023mammut,li2023blip}. Similarly,
CapPa~\cite{tschannen2024image} trains a captioning model that functions as both
a masked and causal decoder, and~\citet{Caron_2024_CVPR} re-purposes a
captioning model for web-scale entity recognition. Different from most previous
approaches, \Ours does not use cross-attention and treats vision and text tokens
symmetrically, similar to large multimodal models (e.g.
LLaVA~\cite{liu2024improvedllava} and MM1~\cite{mckinzie2024mm1}). Additionally,
\Ours incorporates an autoregressive image modeling loss on vision tokens,
further enhancing performance beyond captioning-only methods.

\fi
\section{Conclusion}
This paper introduce \Ours, a family of vision encoders pre-trained with a
multimodal autoregressive objective that reconstructs image patches and
text tokens. This unified objective enables \Ours to excel in diverse tasks,
including image recognition, grounding, and multimodal understanding. The
superior performance of \Ours is attributed to its ability to leverage signals
from all input tokens and patches, facilitating efficient training with fewer
samples compared to other methods. \Ours consistently outperforms or matches
existing self-supervised and vision-language pre-trained models, demonstrating
its strength and versatility as a vision encoder. Additionally, the
straightforward pre-training process of \Ours ensures easy scalability, paving
the way for further advancements in vision model scaling.

\ifarxiv
    \vspace{-1mm}
    \section*{Acknowledgement}
    \vspace{-1mm}
    We thank Shuangfei Zhai, Miguel Bautista, Jason Ramapuram, Chun-Liang Li,
    Seyed Mohsen Moosavi Dezfooli, David Mizrahi, Roman Bachmann, and Jesse
    Allardice for many fruitful discussions. We thank Vaishaal Shankar, Peter
    Grasch, Vasileios Saveris, and Jeff Lai for their support on data collection
    and synthetic captioning. We thank Denise Hui, Dan Busbridge, and Samy
    Bengio for infra and compute support. Finally, we thank  Marco
    Cuturi, James Thornton, Pierre Ablin, Eugene Ndiaye and the whole MLR team
    at Apple for their support throughout the project.

\fi

{
    \small
    \bibliographystyle{ieeenat_fullname}
    \bibliography{main}

\begin{thebibliography}{138}
\providecommand{\natexlab}[1]{#1}
\providecommand{\url}[1]{\texttt{#1}}
\expandafter\ifx\csname urlstyle\endcsname\relax
  \providecommand{\doi}[1]{doi: #1}\else
  \providecommand{\doi}{doi: \begingroup \urlstyle{rm}\Url}\fi

\bibitem[Achiam et~al.(2023)Achiam, Adler, Agarwal, Ahmad, Akkaya, Aleman, Almeida, Altenschmidt, Altman, Anadkat, et~al.]{achiam2023gpt}
Josh Achiam, Steven Adler, Sandhini Agarwal, Lama Ahmad, Ilge Akkaya, Florencia~Leoni Aleman, Diogo Almeida, Janko Altenschmidt, Sam Altman, Shyamal Anadkat, et~al.
\newblock Gpt-4 technical report.
\newblock \emph{arXiv preprint arXiv:2303.08774}, 2023.

\bibitem[Agrawal et~al.(2019)Agrawal, Desai, Wang, Chen, Jain, Johnson, Batra, Parikh, Lee, and Anderson]{agrawal2019nocaps}
Harsh Agrawal, Karan Desai, Yufei Wang, Xinlei Chen, Rishabh Jain, Mark Johnson, Dhruv Batra, Devi Parikh, Stefan Lee, and Peter Anderson.
\newblock Nocaps: Novel object captioning at scale.
\newblock In \emph{ICCV}, 2019.

\bibitem[Alayrac et~al.(2022)Alayrac, Donahue, Luc, Miech, Barr, Hasson, Lenc, Mensch, Millican, Reynolds, Ring, Rutherford, Cabi, Han, Gong, Samangooei, Monteiro, Menick, Borgeaud, Brock, Nematzadeh, Sharifzadeh, Binkowski, Barreira, Vinyals, Zisserman, and Simonyan]{alayrac2022flamingo}
Jean-Baptiste Alayrac, Jeff Donahue, Pauline Luc, Antoine Miech, Iain Barr, Yana Hasson, Karel Lenc, Arthur Mensch, Katie Millican, Malcolm Reynolds, Roman Ring, Eliza Rutherford, Serkan Cabi, Tengda Han, Zhitao Gong, Sina Samangooei, Marianne Monteiro, Jacob Menick, Sebastian Borgeaud, Andrew Brock, Aida Nematzadeh, Sahand Sharifzadeh, Mikolaj Binkowski, Ricardo Barreira, Oriol Vinyals, Andrew Zisserman, and Karen Simonyan.
\newblock Flamingo: a visual language model for few-shot learning.
\newblock In \emph{NeurIPS}, 2022.

\bibitem[Assran et~al.(2023)Assran, Duval, Misra, Bojanowski, Vincent, Rabbat, LeCun, and Ballas]{assran2023self}
Mahmoud Assran, Quentin Duval, Ishan Misra, Piotr Bojanowski, Pascal Vincent, Michael Rabbat, Yann LeCun, and Nicolas Ballas.
\newblock Self-supervised learning from images with a joint-embedding predictive architecture.
\newblock \emph{arXiv preprint arXiv:2301.08243}, 2023.

\bibitem[Bai et~al.(2023{\natexlab{a}})Bai, Bai, Chu, Cui, Dang, Deng, Fan, Ge, Han, Huang, et~al.]{bai2023qwen}
Jinze Bai, Shuai Bai, Yunfei Chu, Zeyu Cui, Kai Dang, Xiaodong Deng, Yang Fan, Wenbin Ge, Yu Han, Fei Huang, et~al.
\newblock Qwen technical report.
\newblock \emph{arXiv preprint arXiv:2309.16609}, 2023{\natexlab{a}}.

\bibitem[Bai et~al.(2023{\natexlab{b}})Bai, Bai, Yang, Wang, Tan, Wang, Lin, Zhou, and Zhou]{bai2023qwenvl}
Jinze Bai, Shuai Bai, Shusheng Yang, Shijie Wang, Sinan Tan, Peng Wang, Junyang Lin, Chang Zhou, and Jingren Zhou.
\newblock Qwen-vl: A frontier large vision-language model with versatile abilities.
\newblock \emph{arXiv preprint arXiv:2308.12966}, 2023{\natexlab{b}}.

\bibitem[Bandi et~al.(2018)Bandi, Geessink, Manson, Van~Dijk, Balkenhol, Hermsen, Bejnordi, Lee, Paeng, Zhong, et~al.]{bandi2018detection}
Peter Bandi, Oscar Geessink, Quirine Manson, Marcory Van~Dijk, Maschenka Balkenhol, Meyke Hermsen, Babak~Ehteshami Bejnordi, Byungjae Lee, Kyunghyun Paeng, Aoxiao Zhong, et~al.
\newblock From detection of individual metastases to classification of lymph node status at the patient level: the camelyon17 challenge.
\newblock \emph{IEEE Transactions on Medical Imaging}, 2018.

\bibitem[Bao et~al.(2022)Bao, Dong, and Wei]{bao2021beit}
Hangbo Bao, Li Dong, and Furu Wei.
\newblock {BEiT}: {B}ert pre-training of image transformers.
\newblock In \emph{ICLR}, 2022.

\bibitem[Beyer et~al.(2023)Beyer, Izmailov, Kolesnikov, Caron, Kornblith, Zhai, Minderer, Tschannen, Alabdulmohsin, and Pavetic]{beyer2023flexivit}
Lucas Beyer, Pavel Izmailov, Alexander Kolesnikov, Mathilde Caron, Simon Kornblith, Xiaohua Zhai, Matthias Minderer, Michael Tschannen, Ibrahim Alabdulmohsin, and Filip Pavetic.
\newblock Flexivit: One model for all patch sizes.
\newblock In \emph{CVPR}, 2023.

\bibitem[Bossard et~al.(2014)Bossard, Guillaumin, and Van~Gool]{bossard14}
Lukas Bossard, Matthieu Guillaumin, and Luc Van~Gool.
\newblock Food-101 -- mining discriminative components with random forests.
\newblock In \emph{ECCV}, 2014.

\bibitem[Box et~al.(2015)Box, Jenkins, Reinsel, and Ljung]{box2015time}
George~EP Box, Gwilym~M Jenkins, Gregory~C Reinsel, and Greta~M Ljung.
\newblock \emph{Time series analysis: forecasting and control}.
\newblock John Wiley \& Sons, 2015.

\bibitem[Brown et~al.(2020)Brown, Mann, Ryder, Subbiah, Kaplan, Dhariwal, Neelakantan, Shyam, Sastry, Askell, et~al.]{brown2020language}
Tom~B Brown, Benjamin Mann, Nick Ryder, Melanie Subbiah, Jared Kaplan, Prafulla Dhariwal, Arvind Neelakantan, Pranav Shyam, Girish Sastry, Amanda Askell, et~al.
\newblock Language models are few-shot learners.
\newblock \emph{preprint arXiv:2005.14165}, 2020.

\bibitem[Byeon et~al.(2022)Byeon, Park, Kim, Lee, Baek, and Kim]{byeon2022coyo}
Minwoo Byeon, Beomhee Park, Haecheon Kim, Sungjun Lee, Woonhyuk Baek, and Saehoon Kim.
\newblock Coyo-700m: Image-text pair dataset, 2022.

\bibitem[Carion et~al.(2020)Carion, Massa, Synnaeve, Usunier, Kirillov, and Zagoruyko]{carion2020end}
Nicolas Carion, Francisco Massa, Gabriel Synnaeve, Nicolas Usunier, Alexander Kirillov, and Sergey Zagoruyko.
\newblock End-to-end object detection with transformers.
\newblock In \emph{ECCV}, 2020.

\bibitem[Caron et~al.(2020)Caron, Misra, Mairal, Goyal, Bojanowski, and Joulin]{caron2020unsupervised}
Mathilde Caron, Ishan Misra, Julien Mairal, Priya Goyal, Piotr Bojanowski, and Armand Joulin.
\newblock Unsupervised learning of visual features by contrasting cluster assignments.
\newblock In \emph{NeurIPS}, 2020.

\bibitem[Caron et~al.(2021)Caron, Touvron, Misra, J\'egou, Mairal, Bojanowski, and Joulin]{caron2021emerging}
Mathilde Caron, Hugo Touvron, Ishan Misra, Herv\'e J\'egou, Julien Mairal, Piotr Bojanowski, and Armand Joulin.
\newblock Emerging properties in self-supervised vision transformers.
\newblock In \emph{ICCV}, 2021.

\bibitem[Caron et~al.(2024)Caron, Iscen, Fathi, and Schmid]{Caron_2024_CVPR}
Mathilde Caron, Ahmet Iscen, Alireza Fathi, and Cordelia Schmid.
\newblock A generative approach for wikipedia-scale visual entity recognition.
\newblock In \emph{CVPR}, 2024.

\bibitem[Chen et~al.(2019)Chen, Wang, Pang, Cao, Xiong, Li, Sun, Feng, Liu, Xu, Zhang, Cheng, Zhu, Cheng, Zhao, Li, Lu, Zhu, Wu, Dai, Wang, Shi, Ouyang, Loy, and Lin]{chen2019mmdetectionopenmmlabdetection}
Kai Chen, Jiaqi Wang, Jiangmiao Pang, Yuhang Cao, Yu Xiong, Xiaoxiao Li, Shuyang Sun, Wansen Feng, Ziwei Liu, Jiarui Xu, Zheng Zhang, Dazhi Cheng, Chenchen Zhu, Tianheng Cheng, Qijie Zhao, Buyu Li, Xin Lu, Rui Zhu, Yue Wu, Jifeng Dai, Jingdong Wang, Jianping Shi, Wanli Ouyang, Chen~Change Loy, and Dahua Lin.
\newblock Mmdetection: Open mmlab detection toolbox and benchmark.
\newblock 2019.

\bibitem[Chen et~al.(2020{\natexlab{a}})Chen, Radford, Child, Wu, Jun, Luan, and Sutskever]{pmlr-v119-chen20s}
Mark Chen, Alec Radford, Rewon Child, Jeffrey Wu, Heewoo Jun, David Luan, and Ilya Sutskever.
\newblock Generative pretraining from pixels.
\newblock In \emph{ICML}, 2020{\natexlab{a}}.

\bibitem[Chen et~al.(2020{\natexlab{b}})Chen, Kornblith, Norouzi, and Hinton]{chen2020simple}
Ting Chen, Simon Kornblith, Mohammad Norouzi, and Geoffrey Hinton.
\newblock A simple framework for contrastive learning of visual representations.
\newblock In \emph{ICML}, 2020{\natexlab{b}}.

\bibitem[Chen et~al.(2015)Chen, Fang, Lin, Vedantam, Gupta, Doll{\'a}r, and Zitnick]{chen2015microsoft}
Xinlei Chen, Hao Fang, Tsung-Yi Lin, Ramakrishna Vedantam, Saurabh Gupta, Piotr Doll{\'a}r, and C~Lawrence Zitnick.
\newblock Microsoft coco captions: Data collection and evaluation server.
\newblock \emph{arXiv preprint arXiv:1504.00325}, 2015.

\bibitem[Chiang et~al.(2023)Chiang, Li, Lin, Sheng, Wu, Zhang, Zheng, Zhuang, Zhuang, Gonzalez, Stoica, and Xing]{vicuna}
Wei-Lin Chiang, Zhuohan Li, Zi Lin, Ying Sheng, Zhanghao Wu, Hao Zhang, Lianmin Zheng, Siyuan Zhuang, Yonghao Zhuang, Joseph~E. Gonzalez, Ion Stoica, and Eric~P. Xing.
\newblock Vicuna: An open-source chatbot impressing gpt-4 with 90\%* chatgpt quality, 2023.

\bibitem[Chowdhery et~al.(2022)Chowdhery, Narang, Devlin, Bosma, Mishra, Roberts, Barham, Chung, Sutton, Gehrmann, et~al.]{chowdhery2022palm}
Aakanksha Chowdhery, Sharan Narang, Jacob Devlin, Maarten Bosma, Gaurav Mishra, Adam Roberts, Paul Barham, Hyung~Won Chung, Charles Sutton, Sebastian Gehrmann, et~al.
\newblock Palm: Scaling language modeling with pathways. arxiv 2022.
\newblock \emph{arXiv preprint arXiv:2204.02311}, 2022.

\bibitem[Christie et~al.(2018)Christie, Fendley, Wilson, and Mukherjee]{christie2018functional}
Gordon Christie, Neil Fendley, James Wilson, and Ryan Mukherjee.
\newblock Functional map of the world.
\newblock In \emph{Proceedings of the IEEE Conference on Computer Vision and Pattern Recognition}, 2018.

\bibitem[Cimpoi et~al.(2014)Cimpoi, Maji, Kokkinos, Mohamed, and Vedaldi]{cimpoi14describing}
M. Cimpoi, S. Maji, I. Kokkinos, S. Mohamed, and A. Vedaldi.
\newblock Describing textures in the wild.
\newblock In \emph{CVPR}, 2014.

\bibitem[Dehghani et~al.(2024)Dehghani, Mustafa, Djolonga, Heek, Minderer, Caron, Steiner, Puigcerver, Geirhos, Alabdulmohsin, et~al.]{dehghani2024patch}
Mostafa Dehghani, Basil Mustafa, Josip Djolonga, Jonathan Heek, Matthias Minderer, Mathilde Caron, Andreas Steiner, Joan Puigcerver, Robert Geirhos, Ibrahim~M Alabdulmohsin, et~al.
\newblock Patch n’pack: Navit, a vision transformer for any aspect ratio and resolution.
\newblock \emph{Advances in Neural Information Processing Systems}, 36, 2024.

\bibitem[Deng et~al.(2009)Deng, Dong, Socher, Li, Li, and Fei-Fei]{deng2009imagenet}
Jia Deng, Wei Dong, Richard Socher, Li-Jia Li, Kai Li, and Li Fei-Fei.
\newblock Imagenet: A large-scale hierarchical image database.
\newblock In \emph{2009 IEEE conference on computer vision and pattern recognition}, 2009.

\bibitem[Desai and Johnson(2021)]{desai2021virtex}
Karan Desai and Justin Johnson.
\newblock Virtex: Learning visual representations from textual annotations.
\newblock In \emph{Proceedings of the IEEE/CVF conference on computer vision and pattern recognition}, 2021.

\bibitem[Doersch et~al.(2015)Doersch, Gupta, and Efros]{doersch2015unsupervised}
Carl Doersch, Abhinav Gupta, and Alexei~A Efros.
\newblock Unsupervised visual representation learning by context prediction.
\newblock In \emph{ICCV}, 2015.

\bibitem[Dosovitskiy et~al.(2021)Dosovitskiy, Beyer, Kolesnikov, Weissenborn, Zhai, Unterthiner, Dehghani, Minderer, Heigold, Gelly, et~al.]{dosovitskiy2020image}
Alexey Dosovitskiy, Lucas Beyer, Alexander Kolesnikov, Dirk Weissenborn, Xiaohua Zhai, Thomas Unterthiner, Mostafa Dehghani, Matthias Minderer, Georg Heigold, Sylvain Gelly, et~al.
\newblock An image is worth 16x16 words: Transformers for image recognition at scale.
\newblock In \emph{ICLR}, 2021.

\bibitem[Dubey et~al.(2024{\natexlab{a}})Dubey, Jauhri, Pandey, Kadian, Al-Dahle, Letman, Mathur, Schelten, Yang, Fan, et~al.]{dubey2024llama}
Abhimanyu Dubey, Abhinav Jauhri, Abhinav Pandey, Abhishek Kadian, Ahmad Al-Dahle, Aiesha Letman, Akhil Mathur, Alan Schelten, Amy Yang, Angela Fan, et~al.
\newblock The llama 3 herd of models.
\newblock \emph{arXiv preprint arXiv:2407.21783}, 2024{\natexlab{a}}.

\bibitem[Dubey et~al.(2024{\natexlab{b}})Dubey, Jauhri, Pandey, Kadian, Al-Dahle, Letman, Mathur, Schelten, Yang, Fan, et~al.]{llama3}
Abhimanyu Dubey, Abhinav Jauhri, Abhinav Pandey, Abhishek Kadian, Ahmad Al-Dahle, Aiesha Letman, Akhil Mathur, Alan Schelten, Amy Yang, Angela Fan, et~al.
\newblock The llama 3 herd of models.
\newblock \emph{arXiv preprint arXiv:2407.21783}, 2024{\natexlab{b}}.

\bibitem[El-Nouby et~al.(2024)El-Nouby, Klein, Zhai, Bautista, Toshev, Shankar, Susskind, and Joulin]{el2024scalable}
Alaaeldin El-Nouby, Michal Klein, Shuangfei Zhai, Miguel~Angel Bautista, Alexander Toshev, Vaishaal Shankar, Joshua~M Susskind, and Armand Joulin.
\newblock Scalable pre-training of large autoregressive image models.
\newblock \emph{arXiv preprint arXiv:2401.08541}, 2024.

\bibitem[Esser et~al.(2021)Esser, Rombach, and Ommer]{esser2021taming}
Patrick Esser, Robin Rombach, and Bjorn Ommer.
\newblock Taming transformers for high-resolution image synthesis.
\newblock In \emph{CVPR}, 2021.

\bibitem[Fang et~al.(2023)Fang, Jose, Jain, Schmidt, Toshev, and Shankar]{fang2023data}
Alex Fang, Albin~Madappally Jose, Amit Jain, Ludwig Schmidt, Alexander Toshev, and Vaishaal Shankar.
\newblock Data filtering networks.
\newblock \emph{arXiv preprint arXiv:2309.17425}, 2023.

\bibitem[Feichtenhofer et~al.(2018)Feichtenhofer, Fan, Malik, and He]{feichtenhofer2018slowfast}
Christoph Feichtenhofer, Haoqi Fan, Jitendra Malik, and Kaiming He.
\newblock Slowfast networks for video recognition. 2019 ieee.
\newblock In \emph{ICCV}, 2018.

\bibitem[Fini et~al.(2023)Fini, Astolfi, Romero-Soriano, Verbeek, and Drozdzal]{fini2023improved}
Enrico Fini, Pietro Astolfi, Adriana Romero-Soriano, Jakob Verbeek, and Michal Drozdzal.
\newblock Improved baselines for vision-language pre-training.
\newblock \emph{arXiv preprint arXiv:2305.08675}, 2023.

\bibitem[Fu et~al.(2023)Fu, Chen, Shen, Qin, Zhang, Lin, Qiu, Lin, Yang, Zheng, et~al.]{fu2023mme}
Chaoyou Fu, Peixian Chen, Yunhang Shen, Yulei Qin, Mengdan Zhang, Xu Lin, Zhenyu Qiu, Wei Lin, Jinrui Yang, Xiawu Zheng, et~al.
\newblock Mme: A comprehensive evaluation benchmark for multimodal large language models.
\newblock \emph{arXiv preprint arXiv:2306.13394}, 2023.

\bibitem[Gidaris et~al.(2018)Gidaris, Singh, and Komodakis]{gidaris2018unsupervised}
Spyros Gidaris, Praveer Singh, and Nikos Komodakis.
\newblock Unsupervised representation learning by predicting image rotations.
\newblock \emph{arXiv preprint arXiv:1803.07728}, 2018.

\bibitem[Goyal et~al.(2017{\natexlab{a}})Goyal, Khot, Summers-Stay, Batra, and Parikh]{goyal2017making}
Yash Goyal, Tejas Khot, Douglas Summers-Stay, Dhruv Batra, and Devi Parikh.
\newblock Making the v in vqa matter: Elevating the role of image understanding in visual question answering.
\newblock In \emph{CVPR}, 2017{\natexlab{a}}.

\bibitem[Goyal et~al.(2017{\natexlab{b}})Goyal, Khot, Summers-Stay, Batra, and Parikh]{goyal2017vqav2}
Yash Goyal, Tejas Khot, Douglas Summers-Stay, Dhruv Batra, and Devi Parikh.
\newblock Making the v in vqa matter: Elevating the role of image understanding in visual question answering.
\newblock In \emph{Proceedings of the IEEE conference on computer vision and pattern recognition}, 2017{\natexlab{b}}.

\bibitem[Grill et~al.(2020)Grill, Strub, Altch{\'e}, Tallec, Richemond, Buchatskaya, Doersch, Avila~Pires, Guo, Gheshlaghi~Azar, et~al.]{grill2020bootstrap}
Jean-Bastien Grill, Florian Strub, Florent Altch{\'e}, Corentin Tallec, Pierre Richemond, Elena Buchatskaya, Carl Doersch, Bernardo Avila~Pires, Zhaohan Guo, Mohammad Gheshlaghi~Azar, et~al.
\newblock Bootstrap your own latent-a new approach to self-supervised learning.
\newblock \emph{NeurIPS}, 2020.

\bibitem[Gupta et~al.(2019)Gupta, Dollár, and Girshick]{gupta2019lvisdatasetlargevocabulary}
Agrim Gupta, Piotr Dollár, and Ross Girshick.
\newblock Lvis: A dataset for large vocabulary instance segmentation, 2019.

\bibitem[Gurari et~al.(2018)Gurari, Li, Stangl, Guo, Lin, Grauman, Luo, and Bigham]{gurari2018vizwiz}
Danna Gurari, Qing Li, Abigale~J Stangl, Anhong Guo, Chi Lin, Kristen Grauman, Jiebo Luo, and Jeffrey~P Bigham.
\newblock Vizwiz grand challenge: Answering visual questions from blind people.
\newblock In \emph{CVPR}, 2018.

\bibitem[He et~al.(2016)He, Zhang, Ren, and Sun]{he2016deep}
Kaiming He, Xiangyu Zhang, Shaoqing Ren, and Jian Sun.
\newblock Deep residual learning for image recognition.
\newblock In \emph{CVPR}, 2016.

\bibitem[He et~al.(2017)He, Gkioxari, Doll{\'a}r, and Girshick]{he2017mask}
Kaiming He, Georgia Gkioxari, Piotr Doll{\'a}r, and Ross Girshick.
\newblock Mask r-cnn.
\newblock In \emph{ICCV}, 2017.

\bibitem[He et~al.(2018)He, Gkioxari, Dollár, and Girshick]{he2018maskrcnn}
Kaiming He, Georgia Gkioxari, Piotr Dollár, and Ross Girshick.
\newblock Mask r-cnn.
\newblock 2018.

\bibitem[He et~al.(2022)He, Chen, Xie, Li, Doll{\'a}r, and Girshick]{he2021masked}
Kaiming He, Xinlei Chen, Saining Xie, Yanghao Li, Piotr Doll{\'a}r, and Ross Girshick.
\newblock Masked autoencoders are scalable vision learners.
\newblock In \emph{CVPR}, 2022.

\bibitem[Helber et~al.(2017)Helber, Bischke, Dengel, and Borth]{helber2017eurosat}
Patrick Helber, Benjamin Bischke, Andreas Dengel, and Damian Borth.
\newblock Eurosat: A novel dataset and deep learning benchmark for land use and land cover classification, 2017.

\bibitem[Hoffmann et~al.(2022)Hoffmann, Borgeaud, Mensch, Buchatskaya, Cai, Rutherford, Casas, Hendricks, Welbl, Clark, et~al.]{hoffmann2022training}
Jordan Hoffmann, Sebastian Borgeaud, Arthur Mensch, Elena Buchatskaya, Trevor Cai, Eliza Rutherford, Diego de~Las Casas, Lisa~Anne Hendricks, Johannes Welbl, Aidan Clark, et~al.
\newblock Training compute-optimal large language models.
\newblock \emph{arXiv preprint arXiv:2203.15556}, 2022.

\bibitem[Hu et~al.(2022)Hu, Gan, Wang, Yang, Liu, Lu, and Wang]{hu2022scaling}
Xiaowei Hu, Zhe Gan, Jianfeng Wang, Zhengyuan Yang, Zicheng Liu, Yumao Lu, and Lijuan Wang.
\newblock Scaling up vision-language pre-training for image captioning.
\newblock In \emph{CVPR}, 2022.

\bibitem[Hudson and Manning(2019{\natexlab{a}})]{gqa2019}
Drew~A Hudson and Christopher~D Manning.
\newblock Gqa: A new dataset for real-world visual reasoning and compositional question answering.
\newblock In \emph{Proceedings of the IEEE/CVF conference on computer vision and pattern recognition}, 2019{\natexlab{a}}.

\bibitem[Hudson and Manning(2019{\natexlab{b}})]{hudson2019gqa}
Drew~A Hudson and Christopher~D Manning.
\newblock Gqa: A new dataset for real-world visual reasoning and compositional question answering.
\newblock In \emph{CVPR}, 2019{\natexlab{b}}.

\bibitem[Jia et~al.(2021)Jia, Yang, Xia, Chen, Parekh, Pham, Le, Sung, Li, and Duerig]{jia2021scaling}
Chao Jia, Yinfei Yang, Ye Xia, Yi-Ting Chen, Zarana Parekh, Hieu Pham, Quoc Le, Yun-Hsuan Sung, Zhen Li, and Tom Duerig.
\newblock Scaling up visual and vision-language representation learning with noisy text supervision.
\newblock In \emph{ICML}, 2021.

\bibitem[Kaplan et~al.(2020)Kaplan, McCandlish, Henighan, Brown, Chess, Child, Gray, Radford, Wu, and Amodei]{kaplan2020scaling}
Jared Kaplan, Sam McCandlish, Tom Henighan, Tom~B Brown, Benjamin Chess, Rewon Child, Scott Gray, Alec Radford, Jeffrey Wu, and Dario Amodei.
\newblock Scaling laws for neural language models.
\newblock \emph{arXiv preprint arXiv:2001.08361}, 2020.

\bibitem[Karpathy and Fei-Fei(2015)]{karpathy2015deep}
Andrej Karpathy and Li Fei-Fei.
\newblock Deep visual-semantic alignments for generating image descriptions.
\newblock In \emph{CVPR}, 2015.

\bibitem[Kazemzadeh et~al.(2014)Kazemzadeh, Ordonez, Matten, and Berg]{kazemzadeh2014referitgame}
Sahar Kazemzadeh, Vicente Ordonez, Mark Matten, and Tamara Berg.
\newblock Referitgame: Referring to objects in photographs of natural scenes.
\newblock In \emph{Proceedings of the 2014 conference on empirical methods in natural language processing (EMNLP)}, 2014.

\bibitem[Kolesnikov et~al.(2020)Kolesnikov, Beyer, Zhai, Puigcerver, Yung, Gelly, and Houlsby]{kolesnikov2020big}
Alexander Kolesnikov, Lucas Beyer, Xiaohua Zhai, Joan Puigcerver, Jessica Yung, Sylvain Gelly, and Neil Houlsby.
\newblock Big transfer (bit): General visual representation learning.
\newblock In \emph{ECCV}, 2020.

\bibitem[Krause et~al.(2013)Krause, Stark, Deng, and Fei-Fei]{KrauseStarkDengFei-Fei_3DRR2013}
Jonathan Krause, Michael Stark, Jia Deng, and Li Fei-Fei.
\newblock 3d object representations for fine-grained categorization.
\newblock In \emph{4th International IEEE Workshop on 3D Representation and Recognition (3dRR-13)}, 2013.

\bibitem[Krizhevsky et~al.(2009)Krizhevsky, Hinton, et~al.]{krizhevsky2009learning}
Alex Krizhevsky, Geoffrey Hinton, et~al.
\newblock Learning multiple layers of features from tiny images.
\newblock 2009.

\bibitem[Krizhevsky et~al.(2012)Krizhevsky, Sutskever, and Hinton]{krizhevsky2012imagenet}
Alex Krizhevsky, Ilya Sutskever, and Geoffrey~E Hinton.
\newblock Imagenet classification with deep convolutional neural networks.
\newblock \emph{NeurIPS}, 2012.

\bibitem[Kuo et~al.(2023)Kuo, Piergiovanni, Kim, Luo, Caine, Li, Ogale, Zhou, Dai, Chen, et~al.]{kuo2023mammut}
Weicheng Kuo, AJ Piergiovanni, Dahun Kim, Xiyang Luo, Ben Caine, Wei Li, Abhijit Ogale, Luowei Zhou, Andrew Dai, Zhifeng Chen, et~al.
\newblock Mammut: A simple architecture for joint learning for multimodal tasks.
\newblock \emph{arXiv preprint arXiv:2303.16839}, 2023.

\bibitem[Lai et~al.(2024)Lai, Saveris, Chen, Chen, Zhang, Zhang, Tebar, Hu, Gan, Grasch, et~al.]{lai2024revisit}
Zhengfeng Lai, Vasileios Saveris, Chen Chen, Hong-You Chen, Haotian Zhang, Bowen Zhang, Juan~Lao Tebar, Wenze Hu, Zhe Gan, Peter Grasch, et~al.
\newblock Revisit large-scale image-caption data in pre-training multimodal foundation models.
\newblock \emph{arXiv preprint arXiv:2410.02740}, 2024.

\bibitem[Li et~al.(2023{\natexlab{a}})Li, Wang, Wang, Ge, Ge, and Shan]{li2023seed}
Bohao Li, Rui Wang, Guangzhi Wang, Yuying Ge, Yixiao Ge, and Ying Shan.
\newblock Seed-bench: Benchmarking multimodal llms with generative comprehension.
\newblock \emph{arXiv preprint arXiv:2307.16125}, 2023{\natexlab{a}}.

\bibitem[Li et~al.(2021)Li, Selvaraju, Gotmare, Joty, Xiong, and Hoi]{li2021align}
Junnan Li, Ramprasaath Selvaraju, Akhilesh Gotmare, Shafiq Joty, Caiming Xiong, and Steven Chu~Hong Hoi.
\newblock Align before fuse: Vision and language representation learning with momentum distillation.
\newblock \emph{NeurIPS}, 2021.

\bibitem[Li et~al.(2022{\natexlab{a}})Li, Li, Xiong, and Hoi]{li2022blip}
Junnan Li, Dongxu Li, Caiming Xiong, and Steven Hoi.
\newblock Blip: Bootstrapping language-image pre-training for unified vision-language understanding and generation.
\newblock In \emph{ICML}, 2022{\natexlab{a}}.

\bibitem[Li et~al.(2023{\natexlab{b}})Li, Li, Savarese, and Hoi]{li2023blip}
Junnan Li, Dongxu Li, Silvio Savarese, and Steven Hoi.
\newblock Blip-2: Bootstrapping language-image pre-training with frozen image encoders and large language models.
\newblock In \emph{ICML}, 2023{\natexlab{b}}.

\bibitem[Li et~al.(2022{\natexlab{b}})Li, Mao, Girshick, and He]{li2022exploring}
Yanghao Li, Hanzi Mao, Ross Girshick, and Kaiming He.
\newblock Exploring plain vision transformer backbones for object detection.
\newblock In \emph{European conference on computer vision}, 2022{\natexlab{b}}.

\bibitem[Li et~al.(2022{\natexlab{c}})Li, Mao, Girshick, and He]{li2022exploringplainvisiontransformer}
Yanghao Li, Hanzi Mao, Ross Girshick, and Kaiming He.
\newblock Exploring plain vision transformer backbones for object detection, 2022{\natexlab{c}}.

\bibitem[Li et~al.(2024)Li, Zhang, Wang, Zhong, Chen, Chu, Liu, and Jia]{li2024mini}
Yanwei Li, Yuechen Zhang, Chengyao Wang, Zhisheng Zhong, Yixin Chen, Ruihang Chu, Shaoteng Liu, and Jiaya Jia.
\newblock Mini-gemini: Mining the potential of multi-modality vision language models.
\newblock \emph{arXiv preprint arXiv:2403.18814}, 2024.

\bibitem[Lin et~al.(2014)Lin, Maire, Belongie, Hays, Perona, Ramanan, Doll{\'a}r, and Zitnick]{coco2014}
Tsung-Yi Lin, Michael Maire, Serge Belongie, James Hays, Pietro Perona, Deva Ramanan, Piotr Doll{\'a}r, and C~Lawrence Zitnick.
\newblock Microsoft coco: Common objects in context.
\newblock In \emph{Computer Vision--ECCV 2014: 13th European Conference, Zurich, Switzerland, September 6-12, 2014, Proceedings, Part V 13}, 2014.

\bibitem[Lin et~al.(2023)Lin, Liu, Zhang, Gao, Qiu, Xiao, Qiu, Lin, Shao, Chen, et~al.]{lin2023sphinx}
Ziyi Lin, Chris Liu, Renrui Zhang, Peng Gao, Longtian Qiu, Han Xiao, Han Qiu, Chen Lin, Wenqi Shao, Keqin Chen, et~al.
\newblock Sphinx: The joint mixing of weights, tasks, and visual embeddings for multi-modal large language models.
\newblock \emph{arXiv preprint arXiv:2311.07575}, 2023.

\bibitem[Liu et~al.(2024{\natexlab{a}})Liu, Li, Li, and Lee]{liu2024improvedllava}
Haotian Liu, Chunyuan Li, Yuheng Li, and Yong~Jae Lee.
\newblock Improved baselines with visual instruction tuning.
\newblock In \emph{CVPR}, 2024{\natexlab{a}}.

\bibitem[Liu et~al.(2024{\natexlab{b}})Liu, Zeng, Ren, Li, Zhang, Yang, Jiang, Li, Yang, Su, Zhu, and Zhang]{liu2024groundingdinomarryingdino}
Shilong Liu, Zhaoyang Zeng, Tianhe Ren, Feng Li, Hao Zhang, Jie Yang, Qing Jiang, Chunyuan Li, Jianwei Yang, Hang Su, Jun Zhu, and Lei Zhang.
\newblock Grounding dino: Marrying dino with grounded pre-training for open-set object detection.
\newblock 2024{\natexlab{b}}.

\bibitem[Loshchilov and Hutter(2017{\natexlab{a}})]{loshchilov2016sgdr}
Ilya Loshchilov and Frank Hutter.
\newblock Sgdr: Stochastic gradient descent with warm restarts.
\newblock In \emph{ICLR}, 2017{\natexlab{a}}.

\bibitem[Loshchilov and Hutter(2017{\natexlab{b}})]{loshchilov2017decoupled}
Ilya Loshchilov and Frank Hutter.
\newblock Decoupled weight decay regularization.
\newblock \emph{arXiv preprint arXiv:1711.05101}, 2017{\natexlab{b}}.

\bibitem[Lu et~al.(2024)Lu, Clark, Lee, Zhang, Khosla, Marten, Hoiem, and Kembhavi]{lu2024unified}
Jiasen Lu, Christopher Clark, Sangho Lee, Zichen Zhang, Savya Khosla, Ryan Marten, Derek Hoiem, and Aniruddha Kembhavi.
\newblock Unified-io 2: Scaling autoregressive multimodal models with vision language audio and action.
\newblock In \emph{CVPR}, 2024.

\bibitem[Lu et~al.(2022)Lu, Mishra, Xia, Qiu, Chang, Zhu, Tafjord, Clark, and Kalyan]{lu2022learnscienceqa}
Pan Lu, Swaroop Mishra, Tanglin Xia, Liang Qiu, Kai-Wei Chang, Song-Chun Zhu, Oyvind Tafjord, Peter Clark, and Ashwin Kalyan.
\newblock Learn to explain: Multimodal reasoning via thought chains for science question answering.
\newblock \emph{Advances in Neural Information Processing Systems}, 2022.

\bibitem[Mao et~al.(2016)Mao, Huang, Toshev, Camburu, Yuille, and Murphy]{mao2016generation}
Junhua Mao, Jonathan Huang, Alexander Toshev, Oana Camburu, Alan~L Yuille, and Kevin Murphy.
\newblock Generation and comprehension of unambiguous object descriptions.
\newblock In \emph{Proceedings of the IEEE conference on computer vision and pattern recognition}, 2016.

\bibitem[Marino et~al.(2019{\natexlab{a}})Marino, Rastegari, Farhadi, and Mottaghi]{marino2019ok}
Kenneth Marino, Mohammad Rastegari, Ali Farhadi, and Roozbeh Mottaghi.
\newblock Ok-vqa: A visual question answering benchmark requiring external knowledge.
\newblock In \emph{CVPR}, 2019{\natexlab{a}}.

\bibitem[Marino et~al.(2019{\natexlab{b}})Marino, Rastegari, Farhadi, and Mottaghi]{okvqa}
Kenneth Marino, Mohammad Rastegari, Ali Farhadi, and Roozbeh Mottaghi.
\newblock Ok-vqa: A visual question answering benchmark requiring external knowledge.
\newblock In \emph{Conference on Computer Vision and Pattern Recognition (CVPR)}, 2019{\natexlab{b}}.

\bibitem[Masry et~al.(2022)Masry, Long, Tan, Joty, and Hoque]{masry2022chartqa}
Ahmed Masry, Do~Xuan Long, Jia~Qing Tan, Shafiq Joty, and Enamul Hoque.
\newblock Chartqa: A benchmark for question answering about charts with visual and logical reasoning.
\newblock \emph{arXiv preprint arXiv:2203.10244}, 2022.

\bibitem[Mathew et~al.(2021)Mathew, Karatzas, and Jawahar]{mathew2021docvqa}
Minesh Mathew, Dimosthenis Karatzas, and CV Jawahar.
\newblock Docvqa: A dataset for vqa on document images.
\newblock In \emph{Proceedings of the IEEE/CVF winter conference on applications of computer vision}, 2021.

\bibitem[Mathew et~al.(2022)Mathew, Bagal, Tito, Karatzas, Valveny, and Jawahar]{mathew2022infographicvqa}
Minesh Mathew, Viraj Bagal, Rub{\`e}n Tito, Dimosthenis Karatzas, Ernest Valveny, and CV Jawahar.
\newblock Infographicvqa.
\newblock In \emph{Proceedings of the IEEE/CVF Winter Conference on Applications of Computer Vision}, 2022.

\bibitem[McKinzie et~al.(2024)McKinzie, Gan, Fauconnier, Dodge, Zhang, Dufter, Shah, Du, Peng, Weers, et~al.]{mckinzie2024mm1}
Brandon McKinzie, Zhe Gan, Jean-Philippe Fauconnier, Sam Dodge, Bowen Zhang, Philipp Dufter, Dhruti Shah, Xianzhi Du, Futang Peng, Floris Weers, et~al.
\newblock Mm1: Methods, analysis \& insights from multimodal llm pre-training.
\newblock \emph{arXiv preprint arXiv:2403.09611}, 2024.

\bibitem[Noroozi and Favaro(2016)]{noroozi2016unsupervised}
Mehdi Noroozi and Paolo Favaro.
\newblock Unsupervised learning of visual representations by solving jigsaw puzzles.
\newblock In \emph{ECCV}, 2016.

\bibitem[Oquab et~al.(2023)Oquab, Darcet, Moutakanni, Vo, Szafraniec, Khalidov, Fernandez, Haziza, Massa, El-Nouby, Howes, Huang, Xu, Sharma, Li, Galuba, Rabbat, Assran, Ballas, Synnaeve, Misra, Jegou, Mairal, Labatut, Joulin, and Bojanowski]{oquab2023dinov2}
Maxime Oquab, Timothée Darcet, Theo Moutakanni, Huy~V. Vo, Marc Szafraniec, Vasil Khalidov, Pierre Fernandez, Daniel Haziza, Francisco Massa, Alaaeldin El-Nouby, Russell Howes, Po-Yao Huang, Hu Xu, Vasu Sharma, Shang-Wen Li, Wojciech Galuba, Mike Rabbat, Mido Assran, Nicolas Ballas, Gabriel Synnaeve, Ishan Misra, Herve Jegou, Julien Mairal, Patrick Labatut, Armand Joulin, and Piotr Bojanowski.
\newblock Dinov2: Learning robust visual features without supervision, 2023.

\bibitem[Parkhi et~al.(2012)Parkhi, Vedaldi, Zisserman, and Jawahar]{parkhi12a}
O.~M. Parkhi, A. Vedaldi, A. Zisserman, and C.~V. Jawahar.
\newblock Cats and dogs.
\newblock In \emph{CVPR}, 2012.

\bibitem[Peng et~al.(2019)Peng, Bai, Xia, Huang, Saenko, and Wang]{peng2019moment}
Xingchao Peng, Qinxun Bai, Xide Xia, Zijun Huang, Kate Saenko, and Bo Wang.
\newblock Moment matching for multi-source domain adaptation.
\newblock In \emph{ICCV}, 2019.

\bibitem[Plummer et~al.(2017)Plummer, Wang, Cervantes, Caicedo, Hockenmaier, and Lazebnik]{flickr30kentities}
Bryan~A. Plummer, Liwei Wang, Christopher~M. Cervantes, Juan~C. Caicedo, Julia Hockenmaier, and Svetlana Lazebnik.
\newblock Flickr30k entities: Collecting region-to-phrase correspondences for richer image-to-sentence models.
\newblock \emph{IJCV}, 2017.

\bibitem[Pouransari et~al.(2024)Pouransari, Li, Chang, Vasu, Koc, Shankar, and Tuzel]{pouransari2024dataset}
Hadi Pouransari, Chun-Liang Li, Jen-Hao~Rick Chang, Pavan Kumar~Anasosalu Vasu, Cem Koc, Vaishaal Shankar, and Oncel Tuzel.
\newblock Dataset decomposition: Faster llm training with variable sequence length curriculum.
\newblock \emph{arXiv preprint arXiv:2405.13226}, 2024.

\bibitem[Radford et~al.(2018)Radford, Narasimhan, Salimans, and Sutskever]{radford2018improving}
Alec Radford, Karthik Narasimhan, Tim Salimans, and Ilya Sutskever.
\newblock Improving language understanding by generative pre-training.
\newblock 2018.

\bibitem[Radford et~al.(2019)Radford, Wu, Child, Luan, Amodei, Sutskever, et~al.]{radford2019language}
Alec Radford, Jeffrey Wu, Rewon Child, David Luan, Dario Amodei, Ilya Sutskever, et~al.
\newblock Language models are unsupervised multitask learners.
\newblock \emph{OpenAI blog}, 2019.

\bibitem[Radford et~al.(2021)Radford, Kim, Hallacy, Ramesh, Goh, Agarwal, Sastry, Askell, Mishkin, Clark, et~al.]{radford2021learning}
Alec Radford, Jong~Wook Kim, Chris Hallacy, Aditya Ramesh, Gabriel Goh, Sandhini Agarwal, Girish Sastry, Amanda Askell, Pamela Mishkin, Jack Clark, et~al.
\newblock Learning transferable visual models from natural language supervision.
\newblock In \emph{ICML}, 2021.

\bibitem[Raffel et~al.(2020)Raffel, Shazeer, Roberts, Lee, Narang, Matena, Zhou, Li, and Liu]{raffel2020exploring}
Colin Raffel, Noam Shazeer, Adam Roberts, Katherine Lee, Sharan Narang, Michael Matena, Yanqi Zhou, Wei Li, and Peter~J Liu.
\newblock Exploring the limits of transfer learning with a unified text-to-text transformer.
\newblock \emph{The Journal of Machine Learning Research}, 21\penalty0 (1), 2020.

\bibitem[Reid et~al.(2024)Reid, Savinov, Teplyashin, Lepikhin, Lillicrap, Alayrac, Soricut, Lazaridou, Firat, Schrittwieser, et~al.]{reid2024gemini}
Machel Reid, Nikolay Savinov, Denis Teplyashin, Dmitry Lepikhin, Timothy Lillicrap, Jean-baptiste Alayrac, Radu Soricut, Angeliki Lazaridou, Orhan Firat, Julian Schrittwieser, et~al.
\newblock Gemini 1.5: Unlocking multimodal understanding across millions of tokens of context.
\newblock \emph{arXiv preprint arXiv:2403.05530}, 2024.

\bibitem[Ridnik et~al.(2021)Ridnik, Ben-Baruch, Noy, and Zelnik-Manor]{ridnik2021imagenet}
Tal Ridnik, Emanuel Ben-Baruch, Asaf Noy, and Lihi Zelnik-Manor.
\newblock Imagenet-21k pretraining for the masses.
\newblock \emph{arXiv preprint arXiv:2104.10972}, 2021.

\bibitem[Rombach et~al.(2022)Rombach, Blattmann, Lorenz, Esser, and Ommer]{rombach2022high}
Robin Rombach, Andreas Blattmann, Dominik Lorenz, Patrick Esser, and Bj{\"o}rn Ommer.
\newblock High-resolution image synthesis with latent diffusion models.
\newblock In \emph{CVPR}, 2022.

\bibitem[Sariyildiz et~al.(2020)Sariyildiz, Perez, and Larlus]{sariyildiz2020learning}
Mert~Bulent Sariyildiz, Julien Perez, and Diane Larlus.
\newblock Learning visual representations with caption annotations.
\newblock In \emph{Computer Vision--ECCV 2020: 16th European Conference, Glasgow, UK, August 23--28, 2020, Proceedings, Part VIII 16}. Springer, 2020.

\bibitem[Schuhmann et~al.(2021)Schuhmann, Vencu, Beaumont, Kaczmarczyk, Mullis, Katta, Coombes, Jitsev, and Komatsuzaki]{schuhmann2021laion}
Christoph Schuhmann, Richard Vencu, Romain Beaumont, Robert Kaczmarczyk, Clayton Mullis, Aarush Katta, Theo Coombes, Jenia Jitsev, and Aran Komatsuzaki.
\newblock {Laion-400m: Open dataset of clip-filtered 400 million image-text pairs}.
\newblock In \emph{NeurIPS Workshop}, 2021.

\bibitem[Shao et~al.(2019)Shao, Li, Zhang, Peng, Yu, Zhang, Li, and Sun]{o365v1}
Shuai Shao, Zeming Li, Tianyuan Zhang, Chao Peng, Gang Yu, Xiangyu Zhang, Jing Li, and Jian Sun.
\newblock Objects365: A large-scale, high-quality dataset for object detection.
\newblock In \emph{ICCV}, 2019.

\bibitem[Shazeer(2020)]{shazeer2020glu}
Noam Shazeer.
\newblock Glu variants improve transformer.
\newblock \emph{arXiv preprint arXiv:2002.05202}, 2020.

\bibitem[Shi et~al.(2024)Shi, Wu, Mao, Wang, and Darrell]{shi2024we}
Baifeng Shi, Ziyang Wu, Maolin Mao, Xin Wang, and Trevor Darrell.
\newblock When do we not need larger vision models?
\newblock \emph{arXiv preprint arXiv:2403.13043}, 2024.

\bibitem[Shukor et~al.(2023)Shukor, Dancette, Rame, and Cord]{shukor2023unival}
Mustafa Shukor, Corentin Dancette, Alexandre Rame, and Matthieu Cord.
\newblock Unival: Unified model for image, video, audio and language tasks.
\newblock \emph{Transactions on Machine Learning Research Journal}, 2023.

\bibitem[Sidorov et~al.(2020)Sidorov, Hu, Rohrbach, and Singh]{sidorov2020textcaps}
Oleksii Sidorov, Ronghang Hu, Marcus Rohrbach, and Amanpreet Singh.
\newblock Textcaps: a dataset for image captioning with reading comprehension.
\newblock In \emph{ECCV}, 2020.

\bibitem[Singh et~al.(2019{\natexlab{a}})Singh, Natarajan, Shah, Jiang, Chen, Batra, Parikh, and Rohrbach]{singh2019textvqa}
Amanpreet Singh, Vivek Natarajan, Meet Shah, Yu Jiang, Xinlei Chen, Dhruv Batra, Devi Parikh, and Marcus Rohrbach.
\newblock Towards vqa models that can read.
\newblock In \emph{Proceedings of the IEEE/CVF conference on computer vision and pattern recognition}, 2019{\natexlab{a}}.

\bibitem[Singh et~al.(2019{\natexlab{b}})Singh, Natarjan, Shah, Jiang, Chen, Parikh, and Rohrbach]{singh2019towards}
Amanpreet Singh, Vivek Natarjan, Meet Shah, Yu Jiang, Xinlei Chen, Devi Parikh, and Marcus Rohrbach.
\newblock Towards vqa models that can read.
\newblock In \emph{CVPR}, 2019{\natexlab{b}}.

\bibitem[Sun et~al.(2017)Sun, Shrivastava, Singh, and Gupta]{sun2017revisiting}
Chen Sun, Abhinav Shrivastava, Saurabh Singh, and Abhinav Gupta.
\newblock Revisiting unreasonable effectiveness of data in deep learning era.
\newblock In \emph{ICCV}, 2017.

\bibitem[Sun et~al.(2023{\natexlab{a}})Sun, Fang, Wu, Wang, and Cao]{sun2023eva}
Quan Sun, Yuxin Fang, Ledell Wu, Xinlong Wang, and Yue Cao.
\newblock Eva-clip: Improved training techniques for clip at scale.
\newblock \emph{arXiv preprint arXiv:2303.15389}, 2023{\natexlab{a}}.

\bibitem[Sun et~al.(2023{\natexlab{b}})Sun, Yu, Cui, Zhang, Zhang, Wang, Gao, Liu, Huang, and Wang]{sun2023generative}
Quan Sun, Qiying Yu, Yufeng Cui, Fan Zhang, Xiaosong Zhang, Yueze Wang, Hongcheng Gao, Jingjing Liu, Tiejun Huang, and Xinlong Wang.
\newblock Generative pretraining in multimodality.
\newblock \emph{arXiv preprint arXiv:2307.05222}, 2023{\natexlab{b}}.

\bibitem[Sun et~al.(2024)Sun, Cui, Zhang, Zhang, Yu, Wang, Rao, Liu, Huang, and Wang]{sun2024generative}
Quan Sun, Yufeng Cui, Xiaosong Zhang, Fan Zhang, Qiying Yu, Yueze Wang, Yongming Rao, Jingjing Liu, Tiejun Huang, and Xinlong Wang.
\newblock Generative multimodal models are in-context learners.
\newblock In \emph{CVPR}, 2024.

\bibitem[Taylor et~al.(2019)Taylor, Earnshaw, Mabey, Victors, and Yosinski]{taylor2019rxrx1}
J. Taylor, B. Earnshaw, B. Mabey, M. Victors, and J. Yosinski.
\newblock Rxrx1: An image set for cellular morphological variation across many experimental batches.
\newblock In \emph{ICLR}, 2019.

\bibitem[Team(2024)]{team2024chameleon}
Chameleon Team.
\newblock Chameleon: Mixed-modal early-fusion foundation models.
\newblock \emph{arXiv preprint arXiv:2405.09818}, 2024.

\bibitem[Tolias et~al.(2015)Tolias, Sicre, and J{\'e}gou]{tolias2015particular}
Giorgos Tolias, Ronan Sicre, and Herv{\'e} J{\'e}gou.
\newblock Particular object retrieval with integral max-pooling of cnn activations.
\newblock \emph{arXiv preprint arXiv:1511.05879}, 2015.

\bibitem[Tong et~al.(2024)Tong, Brown, Wu, Woo, Middepogu, Akula, Yang, Yang, Iyer, Pan, et~al.]{tong2024cambrian}
Shengbang Tong, Ellis Brown, Penghao Wu, Sanghyun Woo, Manoj Middepogu, Sai~Charitha Akula, Jihan Yang, Shusheng Yang, Adithya Iyer, Xichen Pan, et~al.
\newblock Cambrian-1: A fully open, vision-centric exploration of multimodal llms.
\newblock \emph{arXiv:2406.16860}, 2024.

\bibitem[Touvron et~al.(2023{\natexlab{a}})Touvron, Lavril, Izacard, Martinet, Lachaux, Lacroix, Rozi{\`e}re, Goyal, Hambro, Azhar, et~al.]{touvron2023llama}
Hugo Touvron, Thibaut Lavril, Gautier Izacard, Xavier Martinet, Marie-Anne Lachaux, Timoth{\'e}e Lacroix, Baptiste Rozi{\`e}re, Naman Goyal, Eric Hambro, Faisal Azhar, et~al.
\newblock Llama: Open and efficient foundation language models.
\newblock \emph{arXiv preprint arXiv:2302.13971}, 2023{\natexlab{a}}.

\bibitem[Touvron et~al.(2023{\natexlab{b}})Touvron, Martin, Stone, Albert, Almahairi, Babaei, Bashlykov, Batra, Bhargava, Bhosale, et~al.]{touvron2023llama2}
Hugo Touvron, Louis Martin, Kevin Stone, Peter Albert, Amjad Almahairi, Yasmine Babaei, Nikolay Bashlykov, Soumya Batra, Prajjwal Bhargava, Shruti Bhosale, et~al.
\newblock Llama 2: Open foundation and fine-tuned chat models.
\newblock \emph{arXiv preprint arXiv:2307.09288}, 2023{\natexlab{b}}.

\bibitem[Tschannen et~al.(2024)Tschannen, Kumar, Steiner, Zhai, Houlsby, and Beyer]{tschannen2024image}
Michael Tschannen, Manoj Kumar, Andreas Steiner, Xiaohua Zhai, Neil Houlsby, and Lucas Beyer.
\newblock Image captioners are scalable vision learners too.
\newblock \emph{NeurIPS}, 2024.

\bibitem[Van~Horn et~al.(2018)Van~Horn, Mac~Aodha, Song, Cui, Sun, Shepard, Adam, Perona, and Belongie]{van2018inaturalist}
Grant Van~Horn, Oisin Mac~Aodha, Yang Song, Yin Cui, Chen Sun, Alex Shepard, Hartwig Adam, Pietro Perona, and Serge Belongie.
\newblock The inaturalist species classification and detection dataset.
\newblock In \emph{CVPR}, 2018.

\bibitem[Veeling et~al.(2018)Veeling, Linmans, Winkens, Cohen, and Welling]{veeling2018rotation}
Bastiaan~S Veeling, Jasper Linmans, Jim Winkens, Taco Cohen, and Max Welling.
\newblock Rotation equivariant cnns for digital pathology.
\newblock In \emph{Medical Image Computing and Computer Assisted Intervention}, 2018.

\bibitem[Vinyals et~al.(2015)Vinyals, Toshev, Bengio, and Erhan]{vinyals2015show}
Oriol Vinyals, Alexander Toshev, Samy Bengio, and Dumitru Erhan.
\newblock Show and tell: A neural image caption generator.
\newblock In \emph{Proceedings of the IEEE conference on computer vision and pattern recognition}, 2015.

\bibitem[Wang et~al.(2022)Wang, Yang, Men, Lin, Bai, Li, Ma, Zhou, Zhou, and Yang]{wang2022ofa}
Peng Wang, An Yang, Rui Men, Junyang Lin, Shuai Bai, Zhikang Li, Jianxin Ma, Chang Zhou, Jingren Zhou, and Hongxia Yang.
\newblock Ofa: Unifying architectures, tasks, and modalities through a simple sequence-to-sequence learning framework.
\newblock In \emph{International conference on machine learning}, pages 23318--23340. PMLR, 2022.

\bibitem[Wang et~al.(2021)Wang, Yu, Yu, Dai, Tsvetkov, and Cao]{wang2021simvlm}
Zirui Wang, Jiahui Yu, Adams~Wei Yu, Zihang Dai, Yulia Tsvetkov, and Yuan Cao.
\newblock Simvlm: Simple visual language model pretraining with weak supervision.
\newblock \emph{arXiv preprint arXiv:2108.10904}, 2021.

\bibitem[Wu et~al.(2024)Wu, Zhang, Chen, Tang, Li, Fang, Zhu, Xie, Yin, Yi, et~al.]{wu2024vila}
Yecheng Wu, Zhuoyang Zhang, Junyu Chen, Haotian Tang, Dacheng Li, Yunhao Fang, Ligeng Zhu, Enze Xie, Hongxu Yin, Li Yi, et~al.
\newblock Vila-u: a unified foundation model integrating visual understanding and generation.
\newblock \emph{arXiv preprint arXiv:2409.04429}, 2024.

\bibitem[Xie et~al.(2024)Xie, Mao, Bai, Zhang, Wang, Lin, Gu, Chen, Yang, and Shou]{xie2024show}
Jinheng Xie, Weijia Mao, Zechen Bai, David~Junhao Zhang, Weihao Wang, Kevin~Qinghong Lin, Yuchao Gu, Zhijie Chen, Zhenheng Yang, and Mike~Zheng Shou.
\newblock Show-o: One single transformer to unify multimodal understanding and generation.
\newblock \emph{arXiv preprint arXiv:2408.12528}, 2024.

\bibitem[Xu(2015)]{xu2015show}
Kelvin Xu.
\newblock Show, attend and tell: Neural image caption generation with visual attention.
\newblock \emph{arXiv preprint arXiv:1502.03044}, 2015.

\bibitem[Young et~al.(2014)Young, Lai, Hodosh, and Hockenmaier]{flickr30k}
Peter Young, Alice Lai, Micah Hodosh, and Julia Hockenmaier.
\newblock From image descriptions to visual denotations: New similarity metrics for semantic inference over event descriptions.
\newblock \emph{TACL}, 2, 2014.

\bibitem[Yu et~al.(2021)Yu, Li, Koh, Zhang, Pang, Qin, Ku, Xu, Baldridge, and Wu]{Yu2021VectorquantizedIM}
Jiahui Yu, Xin Li, Jing~Yu Koh, Han Zhang, Ruoming Pang, James Qin, Alexander Ku, Yuanzhong Xu, Jason Baldridge, and Yonghui Wu.
\newblock Vector-quantized image modeling with improved vqgan.
\newblock \emph{ArXiv}, 2021.

\bibitem[Yu et~al.(2022)Yu, Wang, Vasudevan, Yeung, Seyedhosseini, and Wu]{yu2022coca}
Jiahui Yu, Zirui Wang, Vijay Vasudevan, Legg Yeung, Mojtaba Seyedhosseini, and Yonghui Wu.
\newblock Coca: Contrastive captioners are image-text foundation models.
\newblock \emph{TMLR}, 2022.

\bibitem[Yu et~al.(2016)Yu, Poirson, Yang, Berg, and Berg]{yu2016modelingcontextreferringexpressions}
Licheng Yu, Patrick Poirson, Shan Yang, Alexander~C. Berg, and Tamara~L. Berg.
\newblock Modeling context in referring expressions, 2016.

\bibitem[Yu et~al.(2023)Yu, Shi, Pasunuru, Muller, Golovneva, Wang, Babu, Tang, Karrer, Sheynin, et~al.]{yu2023scaling}
Lili Yu, Bowen Shi, Ramakanth Pasunuru, Benjamin Muller, Olga Golovneva, Tianlu Wang, Arun Babu, Binh Tang, Brian Karrer, Shelly Sheynin, et~al.
\newblock Scaling autoregressive multi-modal models: Pretraining and instruction tuning.
\newblock \emph{arXiv preprint arXiv:2309.02591}, 2023.

\bibitem[Zhai et~al.(2022)Zhai, Wang, Mustafa, Steiner, Keysers, Kolesnikov, and Beyer]{zhai2022lit}
Xiaohua Zhai, Xiao Wang, Basil Mustafa, Andreas Steiner, Daniel Keysers, Alexander Kolesnikov, and Lucas Beyer.
\newblock Lit: Zero-shot transfer with locked-image text tuning.
\newblock In \emph{CVPR}, 2022.

\bibitem[Zhai et~al.(2023)Zhai, Mustafa, Kolesnikov, and Beyer]{zhai2023sigmoid}
Xiaohua Zhai, Basil Mustafa, Alexander Kolesnikov, and Lucas Beyer.
\newblock Sigmoid loss for language image pre-training.
\newblock In \emph{ICCV}, 2023.

\bibitem[Zhang and Sennrich(2019)]{zhang2019root}
Biao Zhang and Rico Sennrich.
\newblock Root mean square layer normalization.
\newblock \emph{NeurIPS}, 2019.

\bibitem[Zhang et~al.(2016)Zhang, Isola, and Efros]{zhang2016colorful}
Richard Zhang, Phillip Isola, and Alexei~A Efros.
\newblock Colorful image colorization.
\newblock In \emph{ECCV}, 2016.

\bibitem[Zhao et~al.(2024)Zhao, Chen, Xu, Li, Wang, Li, and Huang]{mmgdino2024}
Xiangyu Zhao, Yicheng Chen, Shilin Xu, Xiangtai Li, Xinjiang Wang, Yining Li, and Haian Huang.
\newblock An open and comprehensive pipeline for unified object grounding and detection, 2024.

\bibitem[Zhou et~al.(2022)Zhou, Wei, Wang, Shen, Xie, Yuille, and Kong]{zhou2021ibot}
Jinghao Zhou, Chen Wei, Huiyu Wang, Wei Shen, Cihang Xie, Alan Yuille, and Tao Kong.
\newblock ibot: Image bert pre-training with online tokenizer.
\newblock In \emph{ICLR}, 2022.

\bibitem[Zhu et~al.(2024)Zhu, Chen, Shen, Li, and Elhoseiny]{zhu2024minigpt4}
Deyao Zhu, Jun Chen, Xiaoqian Shen, Xiang Li, and Mohamed Elhoseiny.
\newblock Mini{GPT}-4: Enhancing vision-language understanding with advanced large language models.
\newblock In \emph{ICLR}, 2024.

\end{thebibliography}
}

\clearpage
\appendix

\ifarxiv
\else
    
\fi
\counterwithin{figure}{section}
\counterwithin{table}{section}
\renewcommand\thefigure{\thesection\arabic{figure}}
\renewcommand\thetable{\thesection\arabic{table}}

\section{Hyperparamters}
\label{app:hparams}

\cpar{Pre-training.} We outline the optimization hyperaparmeters and data
augmentations used during \Ours pre-training in~\cref{tab:pretrain_hparams}. For
the captions, we adopt the tokenizer used by SigLIP~\cite{zhai2023sigmoid} and
truncate any text longer than 77 tokens.

\begin{table}[htb]
    \begin{center}
        \centering
        \setlength{\tabcolsep}{4pt}
        \resizebox{\linewidth}{!}{
        \begin{tabular}{l c c c c}
            config &  ViT-L & ViTs-H & ViT-1B & ViT-3B\\
            \shline
            Optimizer & \multicolumn{4}{c}{Fully decoupled AdamW~\cite{loshchilov2017decoupled}} \\ %
            Optimizer Momentum & \multicolumn{4}{c}{$\beta_1=0.9,\beta_2=0.95$} \\
            Peak learning rate & 1e-3 & 8e-4 & 8e-4 & 4e-4 \\
            Minimum Learning rate & \multicolumn{4}{c}{1e-5} \\
            Weight decay & \multicolumn{4}{c}{1e-4} \\
            Batch size & \multicolumn{4}{c}{8192} \\
            Patch size & \multicolumn{4}{c}{(14, 14)} \\
            Gradient clipping & \multicolumn{4}{c}{1.0} \\
            Warmup iterations & \multicolumn{4}{c}{12,500} \\
            Total iterations & \multicolumn{4}{c}{1,500,000} \\
            Learning rate schedule & \multicolumn{4}{c}{cosine decay~\cite{loshchilov2016sgdr}} \\ %
            Augmentations: \\
            \quad {\tt RandomResizedCrop} \\
            \qquad {\tt size} & \multicolumn{4}{c}{224px} \\
            \qquad {\tt scale} & \multicolumn{4}{c}{[0.4, 1.0]} \\
            \qquad {\tt ratio} & \multicolumn{4}{c}{[0.75, 1.33]} \\
            \qquad {\tt interpolation} & \multicolumn{4}{c}{\texttt{Bicubic}} \\
            \quad {\tt RandomHorizontalFlip} & \multicolumn{4}{c}{$p=0.5$} \\
        \end{tabular}}
    \end{center}
    \caption{\textbf{Pre-training hyperparameters} We detail the hyperaparmeters used for pre-training all \Ours variants.}
    \label{tab:pretrain_hparams}
    \end{table}

\cpar{Attentive probing.} The optimization and data augmentations
hyperaparmeters for the attentive probing stage are detailed
in~\cref{tab:probe_hparams}. We use the same set of hyperaparmeters for all
\Ours capacities and the baselines. To ensure a fair comparison, we sweep the
learning rate and weight decay using the ranges detailed
in~\cref{tab:probe_hparams} and report the strongest results for each model.

\begin{table}[htb]
    \begin{center}
        \centering
        \setlength{\tabcolsep}{2pt}
        \resizebox{\linewidth}{!}{
        \begin{tabular}{lrr}
            config & IN-1k & Others \\
            \shline
            Optimizer & \multicolumn{2}{c}{Pytorch's AdamW~\cite{loshchilov2017decoupled}} \\
            Optimizer Momentum & \multicolumn{2}{c}{$\beta_1=0.9,\beta_2=0.999$} \\
            Peak learning rate grid & \multicolumn{2}{c}{[5e-5, 1e-4, 2e-4, 3e-4, 5e-4, 1e-3, 2e-3]} \\
            Minimum Learning rate & \multicolumn{2}{c}{1e-5} \\
            Weight decay & \multicolumn{2}{c}{[0.05, 0.1]} \\
            Batch size & 1024 & 512 \\
            Gradient clipping & \multicolumn{2}{c}{3.0} \\
            Warmup epochs & 5 & 0 \\
            Epochs & \multicolumn{2}{c}{100}  \\
            Learning rate schedule & \multicolumn{2}{c}{cosine decay} \\ %
            Augmentations: \\
            \quad {\tt RandomResizedCrop} \\
            \qquad {\tt size} & \multicolumn{2}{c}{224px} \\
            \qquad {\tt scale} & \multicolumn{2}{c}{[0.08, 1.0]} \\
            \qquad {\tt ratio} & \multicolumn{2}{c}{[0.75, 1.33]} \\
            \qquad {\tt interpolation} & \multicolumn{2}{c}{\texttt{Bicubic}} \\
            \quad {\tt RandomHorizontalFlip} & \multicolumn{2}{c}{$p=0.5$} \\
            \quad {\tt Color Jitter} & \multicolumn{2}{c}{0.3} \\
            \quad {\tt AutoAugment} & \multicolumn{2}{c}{\texttt{rand-m9-mstd0.5-inc1}} \\
        \end{tabular}}
    \end{center}
    \caption{\textbf{Attentive probe hyperparameters.} We detail the
    hyperparameters used during attentive probing of \Ours and the baselines.
    For \Ours and the baselines we sweep over the learning rates and weight
    decay and report the best performance for each model.}

    \label{tab:probe_hparams}
    \end{table}

\section{Image Recognition}
\label{app:image_recognition}
\cpar{Evaluation benchmarks.} In~\cref{tab:recognition}, we evaluate the
recognition performance of  \Ours and other baselines on a diverse set of
benchmarks that encompass fine-grained recognition, medical imaging, satellite
imagery, natural environment imagery, and infographic analysis. We detail the
datasets, the splits and their sizes in~\cref{tab:recognition_benchmarks}.

\begin{table}[htb!]
    \centering
    \setlength{\tabcolsep}{16pt}
    \scalebox{0.8}{
    \begin{tabular}{lrrr}
        Dataset & train & test & classes \\
         \shline
         Imagenet-1k~\citep{deng2009imagenet} & 1,281,167 & 50,000 & 1000 \\
         iNAT-18~\citep{van2018inaturalist}     & 437,513 & 24,426 & 8142 \\
         CIFAR-10~\citep{krizhevsky2009learning}   & 50,000 & 10,000 & 10 \\
         CIFAR-100~\citep{krizhevsky2009learning}  & 50,000 & 10,000 & 100 \\
         Food101~\citep{bossard14}     & 75,750 & 25,250 & 101 \\ 
         DTD~\citep{cimpoi14describing}       & 3,760 & 1,880 & 47 \\
         Pets~\citep{parkhi12a}      & 3,680 & 3,669 & 37 \\
         Cars~\citep{KrauseStarkDengFei-Fei_3DRR2013}  & 8,144 & 8,041 & 196 \\
         Camelyon17~\citep{bandi2018detection}  & 302,436 & 34904 & 2 \\
         PCAM~\citep{veeling2018rotation}      & 262,144 & 32768 & 2 \\
         RxRx1~\citep{taylor2019rxrx1}      & 40,612 & 9854 & 1139 \\
         EuroSAT~\citep{helber2017eurosat}     & 16,200 & 5400 & 10 \\
         fMoW~\citep{christie2018functional}       & 76,863 & 19915 & 62 \\
         Infograph~\citep{peng2019moment} & 36,023 & 15,582 & 345 \\
    \end{tabular}
    } \caption{\textbf{Recognition benchmarks.} We outline the recognition
    benchmarks, the number of train and test images for each dataset, and the number of
    categories.}
    \label{tab:recognition_benchmarks}
\end{table}

\cpar{High-resolution adaptation.}
In Table~\ref{tab:recognition_high_res}, we show the performance of AIMv2 models
with varying image resolutions (224px, 336px, and 448px) across a broad set of
recognition benchmarks. These results extend the main paper, which primarily
focuses on the 224px resolution and the 3B model at 448px. We observe that
scaling both the model capacity and image resolution leads to consistent
improvements across most tasks.

\begin{table*}[t!]
    \centering
    \setlength{\tabcolsep}{11pt}
    \renewcommand{\arraystretch}{1}
    \resizebox{1\linewidth}{!}{
    \begin{tabular}{llccccccccccccccc}
        model & architecture & \small{\rotatebox{90}{IN-1k}} & \small{\rotatebox{90}{iNAT-18}} & \small{\rotatebox{90}{Cifar10}}  & \small{\rotatebox{90}{Cifar100}} & \small{\rotatebox{90}{Food101}} & \small{\rotatebox{90}{DTD}} & \small{\rotatebox{90}{Pets}} & \small{\rotatebox{90}{Cars}} & \small{\rotatebox{90}{CAM17}} & \small{\rotatebox{90}{PCAM}} & \small{\rotatebox{90}{RxRx1}}  & \small{\rotatebox{90}{EuroSAT}} & \small{\rotatebox{90}{fMoW}} & \small{\rotatebox{90}{Infographic}}\\
         \shline
           & ViT-L/14
           &    86.6
           &    76.0
           &    99.1
           &    92.2
           &    95.7
           &    87.9
           &    96.3
           &    96.3
           &    93.7
           &    89.3
           &    5.6
           &    98.4
           &    60.7
           &    69.0
           \\
          & ViT-H/14
          &    87.5
          &    77.9
          &    99.3
          &    93.5
          &    96.3
          &    88.2
          &    96.6
          &    96.4
          &    93.3
          &    89.3
          &    5.8
          &    98.5
          &    62.2
          &    70.4
          \\
          & ViT-1B/14
          &    88.1
          &    79.7
          &    99.4
          &    94.1
          &    96.7
          &    88.4
          &    96.8
          &    96.5
          &    94.2
          &    89.0
          &    6.7
          &    98.8
          &    63.2
          &    71.7
          \\
          \multirow{-4}{*}{\Ours$_{224\text{px}}$} & ViT-3B/14
          &    88.5
          &    81.5
          &    99.5
          &    94.3
          &    96.8
          &    88.9
          &    97.1
          &    96.5
          &    93.5
          &    89.4
          &    7.3
          &    99.0
          &    64.2
          &    72.2
          \\
          \midrule
          & ViT-L/14 & 87.6 & 79.7 & 99.1 & 92.5 & 96.3 & 88.5 & 96.4 & 96.7 & 93.8 & 89.4 & 6.7 & 98.4 & 62.1 & 71.7
          \\
          & ViT-H/14 & 88.2 & 81.0  & 99.3 & 93.6 & 96.6 & 88.8 & 96.8 & 96.4 & 93.3 & 89.4 & 7.2 & 98.7 & 63.9 & 73.4
          \\
          & ViT-1B/14& 88.7 & 82.7  & 99.4 & 93.9 & 97.1 & 88.9 & 96.9 & 96.5 & 94.2 & 89.5 & 8.4 & 98.9 & 65.1 & 73.7
          \\
          \multirow{-4}{*}{\Ours$_{336\text{px}}$} & 
          ViT-3B/14 & 89.2 & 84.4  & 99.5 & 94.4 & 97.2 & 89.3 & 97.2 & 96.6 & 93.2 & 89.3 & 8.8 & 99.0 & 65.7 & 74.0
          \\
          \midrule
          & ViT-L/14 & 87.9 & 81.3 & 99.1 & 92.4 & 96.6 & 88.9 & 96.5 & 96.6 & 94.1 & 89.6 & 7.4 & 98.6 & 62.8 & 72.7
          \\
          & ViT-H/14 & 88.6 & 82.8 & 99.4 & 93.6 & 97.0 & 88.9 & 96.8 & 96.5 & 93.4 & 89.6 & 7.8 & 98.7 & 64.8 & 74.5
          \\
          & ViT-1B/14 & 89.0 & 83.8& 99.4 & 94.1 & 97.2 & 88.9 & 97.1 & 96.6 & 93.5 & 89.9 & 9.2 & 99.1 & 65.9 & 74.4
          \\
          \multirow{-4}{*}{\Ours$_{448\text{px}}$} & ViT-3B/14 &    89.5 & 85.9  & 99.5 & 94.5 & 97.4 & 89.0 & 97.4 & 96.7 & 93.4 & 89.9 & 9.5 & 98.9 & 66.1 & 74.8
    \end{tabular}} \caption{\textbf{Frozen trunk evaluation for recognition
    benchmarks, high resolution \Ours models.} We report the recognition performance of the \Ours high resolution family of
    models when compared to the base 224px models shown in the main paper. All models are evaluated using attentive probing
    with a frozen backbone.}
    \label{tab:recognition_high_res}
    \vspace{-5mm}
\end{table*}

\label{app:high_res}

\section{Multimodal understanding}
\label{app:sft}

\begin{table}[t!]
    \begin{center}
        \centering
        \setlength{\tabcolsep}{8pt}
        \resizebox{\linewidth}{!}{
        \begin{tabular}{lcc}
            config & Llava SFT mixture & Cambrian \\
            \shline
            Optimizer & \multicolumn{2}{c}{Pytorch's AdamW~\cite{loshchilov2017decoupled}} \\
            Optimizer Momentum & \multicolumn{2}{c}{$\beta_1=0.9,\beta_2=0.999$} \\
            Decoder peak learning rate & 1e-5 & 2e-5 \\
            Connector peak learning rate & 8e-5 & 1.6e-4 \\
            Minimum Learning rate & \multicolumn{2}{c}{0} \\
            Weight decay & \multicolumn{2}{c}{0.01} \\
            Batch size & 128 & 512 \\
            Gradient clipping & \multicolumn{2}{c}{1.0} \\
            Warmup iterations & 250 & 700 \\
            iterations & 5000 & 14,000  \\
            Learning rate schedule & \multicolumn{2}{c}{cosine decay} \\ %
            Transformations & \multicolumn{2}{c}{{\tt [PadToSquare, Resize]}} \\
        \end{tabular}}
    \end{center}
    \caption{\textbf{Instruction tuning hyperaparmeters.} We detail the
    hyperparameters of the instruction tuning stage, both for the Llava SFT
    mixture~\cite{liu2024improvedllava} and Cambrian~\cite{tong2024cambrian}.}
    \label{tab:mm_hparams}
    \end{table}

\subsection{Instruction Tuning Setup}

\cpar{Evaluation benchmarks.} We list the multimodal benchmarks we use for
assessing the performance of our models and the baselines
in~\cref{tab:multimodal_benchmarks}, together with the splits, prompts, and  evaluation metric utilized for each dataset.

\cpar{Hyperparamters.} The hyperaparmeters used for the instruction tuning stage
are detailed in~\cref{tab:mm_hparams}. We use the same hyperaparmeters for all
language decoders, \Ours capacities, and the baselines.

\subsection{Additional Results}

\cpar{Instruction tuning with Cambrian.} 
Table~\ref{tab:multimodal_results_appendix} evaluates AIMv2, fine-tuned on
Cambrian, across different resolutions using a tiling strategy. Unlike the main
paper, which uses Llava SFT, Cambrian offers a less in-domain data mix and
achieves stronger results on text-rich benchmarks. Starting with a base
resolution of 336px (matching the encoder's pretraining resolution), higher
resolutions (672px and 1008px) are obtained with tiling; by splitting high-resolution images
into 2×2 and 3×3 grids. AIMv2 paired with tiling shows consistent improvements
on text-rich benchmarks such as InfoVQA, ChartQA, DocVQA, and TextVQA. However,
on benchmarks like COCO, NoCaps, TextCaps, and MME$_p$, no significant gains are
observed with increased resolution.

\begin{table}[htb!]
    \centering
    \setlength{\tabcolsep}{1pt} %
    \scalebox{0.72}{ %
    \begin{tabular}{lccc}
        Benchmark & Split & Prompt & Evaluation Metric \\
        \shline
         VQAv2 ~\citep{goyal2017vqav2} & Val & \multirow{8}{*}{\parbox{4cm}{\textit{\small Answer the question using a single word or phrase.}}} & Accuracy \\
         GQA ~\citep{gqa2019} & Val & & Accuracy \\
         OKVQA ~\citep{okvqa} & Val & & Accuracy \\
         TextVQA ~\citep{singh2019textvqa} & Val & & Accuracy \\
         DocVQA ~\citep{mathew2021docvqa} & Test & & ANLS \\
         InfoVQA ~\citep{mathew2022infographicvqa} & Test & & ANLS \\
         ChartQA ~\citep{masry2022chartqa} & Test & & Relaxed accuracy \\
         SEED ~\citep{li2023seed} & Test (image split) & & Accuracy \\
        \midrule
         ScienceQA ~\citep{lu2022learnscienceqa} & Test (image split) & \multirow{2}{*}{\parbox{4cm}{\textit{\small Answer with the option’s letter from the given choices directly.}}}  & Accuracy \\
         MME ~\citep{fu2023mme} & Test (image split) & & Accuracy \\

        \midrule 
         COCO ~\citep{coco2014} & Val & \multirow{3}{*}{\parbox{4cm}{\textit{\small Provide a one-sentence caption for the provided image.}}} & CIDEr \\
         TextCaps ~\citep{sidorov2020textcaps} & Val & & CIDEr \\
         NoCaps ~\citep{agrawal2019nocaps} & Val & & CIDEr \\
    \end{tabular}
    } \caption{\textbf{Multimodal benchmarks.} We provide the list of benchmarks
    used during the multimodal evaluation including the reference, split, prompt,
    and the evaluation metric.}
    \label{tab:multimodal_benchmarks}
    \vspace{5mm}
\end{table}

\cpar{Instruction tuning with DCLM-1B decoder.} In~\cref{fig:mm_barcharts_dclm}, we
present the same comparison between OAI CLIP, SigLIP, and \Ours as in the main
paper, but this time using the Llava SFT mixture paired with a DCLM 1B decoder.
These results demonstrate that \Ours consistently outperforms the baselines,
regardless of the decoder's capacity. Notably, in the practical setting of a
small decoder, \Ours maintains its position as the preferred choice for
multimodal understanding tasks.

\subsection{Qualitative Results}
\begin{figure*}[t!]
    \includegraphics[width=\linewidth]{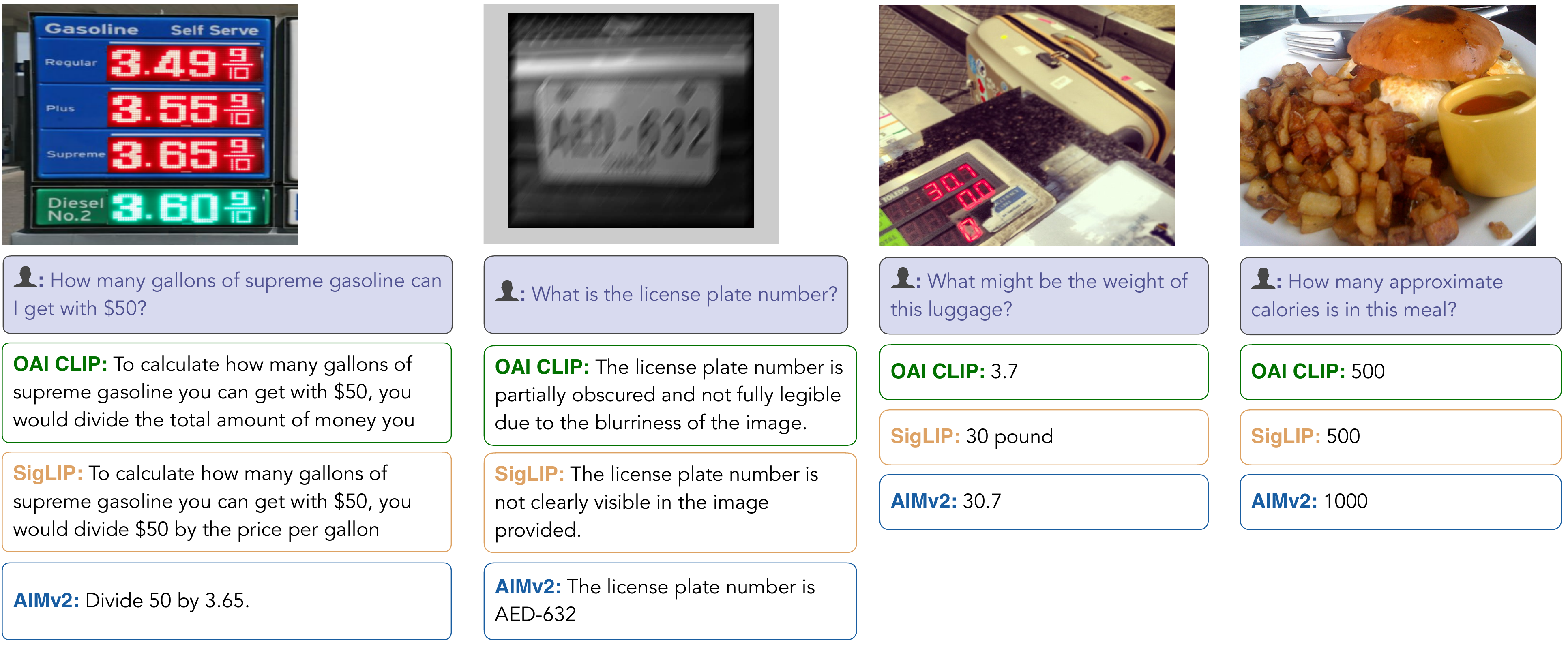}
    \vspace{-3mm}\caption{\textbf{Qualitative comparison of AIMv2, SigLIP, and OAI CLIP on multimodal tasks after instruction tuning on Cambrian.} AIMv2 demonstrates superior performance in both text-rich (e.g. extracting relevant information or reading text in cluttered scenes) and knowledge-based scenarios (e.g., estimating caloric content), showcasing its ability to focus on relevant information, accurately localize text, and provide contextually appropriate answers. }
    \label{fig:qualitative}
\end{figure*}

The qualitative results in~\cref{fig:qualitative} highlight AIMv2's strengths on
multimodal evaluations compared to SigLIP~\cite{zhai2023sigmoid} and OAI
CLIP~\cite{radford2021learning} after instruction tuning on Cambrian. In the
first three examples, AIMv2 excels in text-rich tasks by correctly localizing
and extracting the relevant textual information. For instance, in the example on
the left, AIMv2 is able to identify the correct value for ``supreme gasoline''
and outputs the correct operation for finding the solution (``Divide 50 by
3.65''). This contrasts with OAI CLIP and SigLIP, which provide generic and
incomplete answers that fail to focus on the relevant information. Similarly,
AIMv2 successfully identifies the license plate number (``AED-632'') in a blurry
image, demonstrating robust localization and reading capabilities in challenging
settings. In the luggage example, AIMv2 accurately reads the weight (``30.7''),
despite the presence of multiple distracting objects in the image, while the
other models  make mistakes. Finally, in the calorie estimation example, AIMv2
provides a more plausible response (``1000 calories'') based on its knowledge,
whereas SigLIP and OAI CLIP offer less contextually plausible answers.

\begin{table*}[h!]
    \centering
    \setlength{\tabcolsep}{5pt}
    \renewcommand{\arraystretch}{1}
    \resizebox{1\linewidth}{!}{
    \begin{tabular}{lccccccccccccccc}
        data mix & decoder & resolution & \small{VQAv2} & \small{GQA} & \small{OKVQA}  & \small{TextVQA} & \small{DocVQA} & \small{InfoVQA} & \small{ChartQA} & \small{ScienceQA} & \small{COCO} & \small{TextCaps} & \small{NoCaps} & \small{MME$_\text{p}$}\\
        \shline
        Cambrian & Llama 3.0 & 336px
        & 75.5
        & 71.5
        & 61.1
        & 58.3
        & 50.2
        & 35.1
        & 51.7
        & 78.7
        & 95.5
        & 82.3
        & 98.1
        & 1594        
         \\
        Cambrian & Llama 3.0 & 672px
        & 77.5
        & 72.8
        & 62.0
        & 69.1
        & 76.4
        & 48.3
        & 64.7
        & 79.4
        & 92.6
        & 80.6
        & 95.4
        & 1482
        \\
        Cambrian & Llama 3.0 & 1008px & 77.7 &
        73.2 & 62.0 & 72.2 & 79.2 & 53.5   &
        65.1 & 81.6 & 93.7 & 81.6 &
        97.6 & 1507 \end{tabular}}
        \caption{\textbf{Additional multimodal evaluations.} We compare the performance of \Ours with different SFT data mixtures (Llava~\cite{liu2024improvedllava} and Cambrian~\cite{tong2024cambrian}), and resolutions (336px, 672px and 1008px).}
\label{tab:multimodal_results_appendix}
\end{table*}
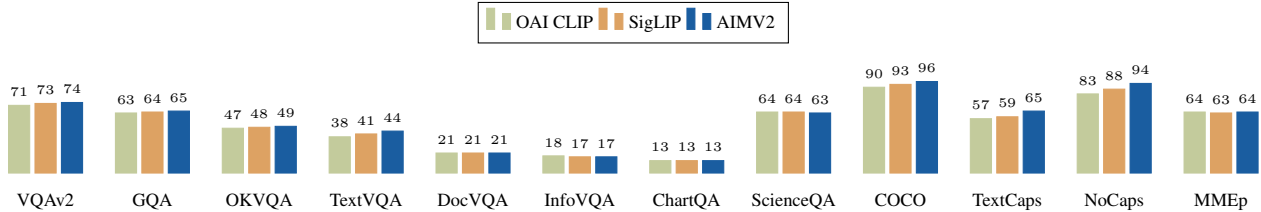
\begin{figure*}[t]
\hspace{-5mm}\subfloat[\textbf{\scriptsize{DCLM 1B + Llava SFT mixture}}]{
        \definecolor{CustomA}{HTML}{879838}
\definecolor{CustomB}{HTML}{DDA263}
\definecolor{CustomC}{HTML}{004B95}

\begin{tikzpicture}
    \begin{axis}[
        ybar,
        bar width=8pt,
        width=\textwidth,
        height=1.15in,
        width=1.1\linewidth,
        ymin=0,
        symbolic x coords={VQAv2, GQA, OKVQA, TextVQA, DocVQA, InfoVQA, ChartQA, ScienceQA, COCO, TextCaps, NoCaps,MMEp},
        yticklabel style={font=\scriptsize},
        xticklabel style={font=\scriptsize},
        ylabel style={align=center, font=\scriptsize, yshift=-0.5ex},
        xlabel style={font=\scriptsize},
        nodes near coords style={font=\tiny, text=black},
        cycle list={{CustomA!50},{CustomB},{CustomC!90}},
        xtick=data,
        nodes near coords,
        legend style={at={(0.5,1.7)}, anchor=north,legend columns=-1, font=\scriptsize},
        enlarge x limits={abs=1cm},
        axis line style={draw=none},
        tick style={draw=none}, 
        ytick=\empty, 
    ]

    \addplot+[fill=CustomA!50] coordinates {
        (VQAv2, 71)
        (GQA, 63)
        (OKVQA, 47)
        (TextVQA, 38)
        (DocVQA, 21)
        (InfoVQA, 18)
        (ChartQA, 13)
        (ScienceQA, 64)
        (COCO,90)
        (TextCaps, 57)
        (NoCaps, 83)
        (MMEp, 64)
    };
    \addplot+[fill=CustomB] coordinates {
        (VQAv2, 73)
        (GQA, 64)
        (OKVQA, 48)
        (TextVQA, 41)
        (DocVQA, 21)
        (InfoVQA, 17)
        (ChartQA, 13)
        (ScienceQA, 64)
        (COCO,93)
        (TextCaps, 59)
        (NoCaps, 88)
        (MMEp, 63)
    };
    \addplot+[fill=CustomC!90] coordinates {
        (VQAv2, 74)
        (GQA, 65)
        (OKVQA, 49)
        (TextVQA, 44)
        (DocVQA, 21)
        (InfoVQA, 17)
        (ChartQA, 13)
        (ScienceQA, 63)
        (COCO,96)
        (TextCaps, 65)
        (NoCaps, 94)
        (MMEp, 64)
    };
    \legend{OAI CLIP, SigLIP, AIMV2}
    \end{axis}
\end{tikzpicture}
    }
    \caption{\textbf{Instruction with a small decoder (DCLM).} Performance comparison of OAI CLIP, SigLIP, and AIMv2 across 12 multimodal benchmarks using the Llava SFT mixture paired with a DCLM 1B decoder. \Ours exhibits superior performance across most benchmarks, even with the constrained capacity of a small decoder.}
    \label{fig:mm_barcharts_dclm}
\end{figure*}

\section{Detection, Segmentation and Grounding}
\label{app:grounding}
    
\subsection{Open Vocabulary Detection and Grounding}

\begin{table}[tb]
    \vspace{-3mm}
    \centering
    \setlength{\tabcolsep}{6pt}
    \renewcommand{\arraystretch}{1}
    \resizebox{1\linewidth}{!}{
    \begin{tabular}{lccccccccc}
     & \multicolumn{4}{c}{COCO} && \multicolumn{4}{c}{LVIS Val} 
    \\
         \textbf{Model} &  $AP_{all}$ & $AP_{s}$ & $AP_{m}$ & $AP_{l}$ && $AP_{all}$ & $AP_r$ & $AP_c$ &
         $AP_f$ \\
          \midrule OpenAI CLIP & 59.1 & 43.5 & 63.5 & 74.8 && \underline{31.0} &
          17.6 & \textbf{27.2} & 41.2 \\
          DFN-CLIP & 59.8 & \underline{44.0} & 63.8 & 75.3 && 30.7 &
          17.2 & 26.4 & \underline{41.5} \\
          SigLIP & 58.8 & 41.7 & 62.8 & \underline{75.7} && 30.5 & 16.5 & 26.5 & 41.1 \\
          DINOv2 & \underline{60.1} & 43.7 & \underline{64.2} & \textbf{75.8} && 30.8 & \textbf{18.5} & 26.1 & 41.4 \\
           \Ours & \textbf{60.2} & \textbf{44.5} & \textbf{64.3} & 75.4 && \textbf{31.6} & \underline{18.0} & \underline{27.0} & \textbf{42.8} \\
     \end{tabular}} \caption{\textbf{Performance on OVD Benchmarks.} We report the performance on mean average precision (AP) for COCO and LVIS. For COCO, we also report AP for the \textit{small}, \textit{medium}, and \textit{large} subsets, while for LVIS, we report on \textit{rare}, \textit{medium}, and \textit{frequent} subsets.}
    \label{tab:grounding_full_OVD}
\end{table}

\begin{table}[tb]
    \vspace{-3mm}
    \centering
    \setlength{\tabcolsep}{6pt}
    \renewcommand{\arraystretch}{1}
    \resizebox{1\linewidth}{!}{
    \begin{tabular}{lcccccc}
     & Window & COCO & LVIS Val & RefCOCO & RefCOCO+ & RefCOCOg
    \\
         \textbf{Model} & Size &  $AP_{all}$ & $AP_{all}$ & Val P@1 & Val P@1 & Val P@1 \\
          \midrule 
          DINOv2 & \multirow{2}{*}{16} & 60.1 & 30.8 & 92.2 & 85.9 & \textbf{89.1} \\
           \Ours & & \textbf{60.2} & \textbf{31.6} & 
           \textbf{92.6} & \textbf{86.3} & 88.9 \\
          \midrule DINOv2 &  & 59.6 & 29.6 & 92.1 & 85.0 & 88.7 \\
           \Ours & \multirow{-2}{*}{24} & \textbf{59.8} & \textbf{31.2} & \textbf{92.3} & \textbf{85.8} & \textbf{89.1} \\
          \midrule DINOv2 & & 60.2 & 30.7 & \textbf{92.5} & 86.1 & \textbf{89.5} \\
           \Ours & \multirow{-2}{*}{32} & \textbf{60.3} & \textbf{32.9} & \textbf{92.5} & \textbf{86.3} & 88.9 \\
          \midrule DINOv2 & 37 & 60.2 & 31.1 & 92.2 & 85.9 & 88.4 \\
     \end{tabular}} \caption{\textbf{Ablation across window sizes.} We report the performance on mean average precision
        (AP) for COCO and LVIS. For RefCOCO* we report Precision @1 on the respective validation splits.}
    \label{tab:grounding_full_WSA}
\end{table}

\cpar{Performance on Small Objects.}
In~\cref{tab:grounding_full_OVD} we report the breakdowns of COCO between
classes that are either \textit{small}, \textit{medium}, or \textit{large}. We
can observe that \Ours consistently outperforms on the \textit{small} classes by
+0.5 AP, compared to DFN-CLIP, the second best performing model in that
breakdown. This is further emphasized by the results reported on LVIS val, as
objects in LVIS are more likely to be small. There we observe an improvement of
+1.3 AP on the \textit{frequent} subset against DFN-CLIP. \\

\cpar{Window Size Ablation.}
Due to varying input resolutions and feature map sizes used during pre-training,
we ablate the effect of window size \cite{li2022exploring} for \Ours and DINOv2
in~\cref{tab:grounding_full_WSA}. For \Ours the input image resolution is scaled
during pre-training such that the feature map size matches the window size
during finetuning, while for DINOv2 the window size is fixed to match \Ours. For
comparison we also add DINOv2 trained with a window size of 37, which matches
its pre-training feature map size. Across the window sizes, \Ours outperforms
DINOv2 across all OVD and for two out of three referring comprehension
benchmarks. When comparing our best performing \Ours with the best performing
DINOv2 across all benchmarks, we observe that \Ours strongly outperforms on LVIS
Val while outperforming on all except one benchmark against DINOv2.

\subsection{Detection and Segmentation via ViTDet Mask-RCNN}
\label{app:cvd_segm}

\begin{table}[tb]
    \vspace{-3mm}
    \centering
    \setlength{\tabcolsep}{6pt}
    \renewcommand{\arraystretch}{1}
    \resizebox{1\linewidth}{!}{
    \begin{tabular}{lcccc c cccc}
    & \multicolumn{4}{c}{detection mAP50:95} && \multicolumn{4}{c}{mask mAP50:95} \\
    \cline{2-5}
    \cline{7-10}
        \textbf{Model} & AP$_{all}$ & AP$_s$ & AP$_m$ & AP$_l$ && AP$_{all}$ & AP$_s$ & AP$_m$ & AP$_l$ \\
        \shline
        OAI CLIP & 53.6 & 37.2 & 58.5 & 69.2 && \underline{46.7} & 26.6 & 50.9 & 66.2 \\ %
        DFN-CLIP & 53.4 & 37.1 & 58.3 & 69.3 && 46.2 & 26.4 & 50.8 & 66.4 \\ %
        SigLIP & 53.3 & 37.2 & 57.6 & 69.7 && 46.6 & \underline{27.1} & 50.5 & 66.3 \\ %
        DINOv2 & \textbf{55.5} & \textbf{39.5} & \textbf{59.9} & \textbf{70.6} && \textbf{48.3} &  \textbf{29.4} & \textbf{52.3} & \textbf{67.4} \\ %
        \Ours & \underline{54.0} & \underline{37.4} & \underline{58.8} & \underline{70.0} && \underline{46.7} & 26.7 & \underline{51.1} & \underline{66.5} \\ %
    \end{tabular}
     }
    \caption{\textbf{COCO17 detection and segmentation benchmarks.} We report overall detection and segmentation scores along with the \textit{small}, \textit{medium}, and \textit{large} subset breakdowns.} 
    \label{tab:coco_full_cvd_seg}
\end{table}

To compare vision only capabilities of the encoders we incorporate them into a
Mask-RCNN\cite{he2018maskrcnn} detection model as backbones by utilizing a
ViTDet formulation to accommodate for high resolution (1024) detector training /
testing input size. We ensure that ViTDet~\cite{li2022exploring} backbone
forward pass outputs match the respective ViT-L implementations before the
training. We utilize the same set of hyperparameters for training all compared
detectors: consistent windowed attention size (16) ensuring comparable compute,
AdamW optimizer, cosine decay learning rate schedule, layer-wise learning rate,
and weight decay. All detectors are fine-tuned on coco17 train split for 100
epochs with a global batch size of 64 following the default recipe from
MMDetection~\cite{chen2019mmdetectionopenmmlabdetection}. We report results from
the coco17-val split in~\cref{tab:coco_full_cvd_seg}. \Ours consistently
outperforms encoders pre-trained on contrastive objectives, falling slightly
behind DINOv2 which provides the strongest performance.

\end{document}